\documentclass[10pt,journal,compsoc]{IEEEtran}

\ifCLASSOPTIONcompsoc
  \usepackage[nocompress]{cite}
\else
  \usepackage{cite}
\fi

\ifCLASSINFOpdf
   \usepackage[pdftex]{graphicx}
\else
   \usepackage[dvips]{graphicx}
\fi

\usepackage{amsmath}
\usepackage{times}
\usepackage{epsfig}
\usepackage{amssymb}
\usepackage{caption}
\usepackage[normalem]{ulem}
\usepackage{multirow}
\usepackage{booktabs}
\usepackage{url}
\usepackage{pifont}

\usepackage[table]{xcolor}
\definecolor{lightgray}{gray}{0.9}
\definecolor{lightblue}{rgb}{0.93,0.95,1.0}
\definecolor{darkgreen}{rgb}{0.0,0.6,0.0}
\definecolor{blue}{rgb}{0, 0, 1}
\usepackage{soul}

\newcommand{\minisection}[1]{\vspace{2mm}\noindent{\textbf{#1}.}}
\newenvironment{tight_itemize}{
\begin{itemize}
  \setlength{\topsep}{0pt}
  \setlength{\itemsep}{2pt}
  \setlength{\parskip}{0pt}
  \setlength{\parsep}{0pt}
}{\end{itemize}}
\usepackage{pifont}

\begin{document}
\title{TextStyleBrush: Transfer of Text Aesthetics from a Single Example}

\author{Praveen Krishnan,
        Rama Kovvuri,
        Guan Pang,
        Boris Vassilev,
        Tal Hassner\\
        Facebook AI

{\small 
\{pkrishnan,ramakovvuri,gpang,borisva,thassner\}@fb.com}
}

\markboth{Journal of \LaTeX\ Class Files,~Vol.~14, No.~8, August~2015}%
{Shell \MakeLowercase{\textit{et al.}}: Bare Demo of IEEEtran.cls for Computer Society Journals}


\IEEEtitleabstractindextext{%
\begin{abstract}
We present a novel approach for disentangling the content of a text image from all aspects of its appearance. The appearance representation we derive can then be applied to new content, for one-shot transfer of the source style to new content. We learn this disentanglement in a self-supervised manner. Our method processes entire word boxes, without requiring segmentation of text from background, per-character processing, or making assumptions on string lengths. We show results in different text domains which were previously handled by specialized methods, e.g., scene text, handwritten text. To these ends, we make a number of technical contributions: (1) We disentangle the style and content of a textual image into a non-parametric, fixed-dimensional vector. (2) We propose a novel approach inspired by StyleGAN but conditioned over the example style at different resolution and content. (3) We present novel self-supervised training criteria which preserve both source style and target content using a pre-trained font classifier and text recognizer. Finally, (4) we also introduce Imgur5K, a new challenging dataset for handwritten word images.  We offer numerous qualitative photo-realistic results of our method. We further show that our method surpasses previous work in quantitative tests on scene text and handwriting datasets, as well as in a user study.


\end{abstract}

}

\maketitle

\IEEEdisplaynontitleabstractindextext

%
\IEEEpeerreviewmaketitle

\IEEEraisesectionheading{\section{Introduction}\label{sec:introduction}}
\label{sec:intro}

\begin{figure}[t]
    \centering
    \includegraphics[width=0.5\textwidth]{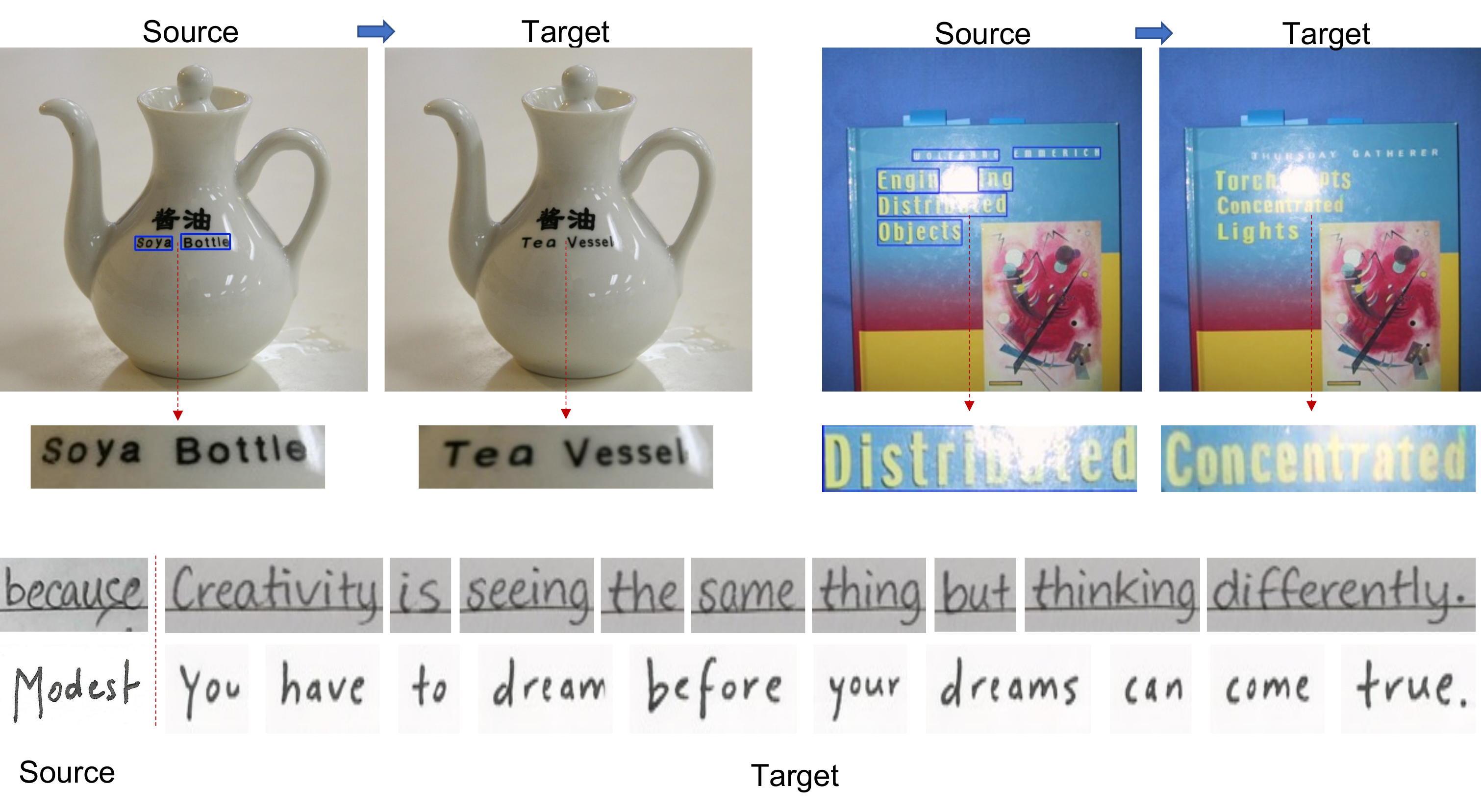}
    \caption{Text generation/editing using TextStyleBrush. (Top) Detected words highlighted in blue boxes edited by changing their content and shown in the target image. (Bottom) Given the source handwritten word image (left), we generate a target sentence image mimicking the same handwritten style (right).}
    \label{fig:teaser}
\end{figure}

\IEEEPARstart{T}{he} term {\em text aesthetics} refers to all aspects of how a text appears, including typography (for printed text~\cite{bringhurst2004elements}), calligraphy (for stylized handwriting~\cite{mediavilla1996calligraphy}), spatial transformations and deformations, and even background clutter and image noise. For simplicity, we refer to all these elements simply as the text {\em Style}. We present a self-supervised method for learning how to disentangle text image style vs. content and one-shot style transfer. We thus enable functionality similar to that of style brush tools in standard word processors, but for text images: Applying the visual style of a source text to new text content (Fig.~\ref{fig:teaser}).

This task is ambitious. Every person has one or more unique handwriting styles which can change over time. To these already limitless text style variations, add all stylized and designer text appearances, used in product logos, street signs~\cite{long2021scene,neumann2012real,wang2011end}, historical manuscripts~\cite{hassner2012computation,hassner2014digital}, and many others, to get an idea of the full range of text style diversity. Moreover, it is unreasonable to expect text images used as style samples to provide example appearances for the entire alphabet (including digits). Transferring styles between strings with different characters therefore implies hallucinating plausible appearances of unseen characters in the input style.

We aim at a general approach, which can be applied to text, irrespective of its domain (e.g., scene text, handwritten). We cannot, therefore, make domain-specific assumptions, such as that the input text can be neatly segmented from its background~\cite{WuZLHLDB19, YangHL20}, that we can process each character separately~\cite{RoyBG020, Yang0WG19, AzadiFKWSD18}, or restrict it to a particular domain~\cite{FogelACML20, davis2020text} (such as handwriting).

We describe {\em text style brush} (TSB) for text-style disentanglement and transfer. Our method is self-supervised and trained on text samples from real photos, without ground truth style labels. Given a detected text box containing a source style, our method extracts an opaque {\em latent style representation}: We do not define this representation using semantic parameters (e.g., typeface, color encoding, spatial transformation). Instead, we optimise our representation to allow photo-realistic rendering of new content in the source style. This transfer is applied in a {\em one-shot} manner, using only the single source sample to learn the desired text style. We do not segment text into individual characters, or assume that the source style and new content share the same length. Soft segmentation masks are learned, however, in a self-supervised manner, in order to improve the quality of foreground (text) produced by our generator.

To enable these novel capabilities, we make the following technical contributions.
\begin{tight_itemize}
\item We propose style and content encoders for one-shot disentanglement of a source style image. 

\item We describe a novel generator, conditioned on content and multi-scale style which generates the target style image along with the foreground mask.

\item Our method is self-supervised, learning to preserve both source style and target content using novel loss functions.

\item We release a new large-scale handwritten \textit{in-the-wild} dataset, Imgur5K, containing challenging real world handwritten samples from nearly 5K writers. 
\end{tight_itemize}

The paper and supplementary material provide numerous qualitative examples of scene and handwritten text style transfers (e.g., Fig.~\ref{fig:teaser}). We further report quantitative results, demonstrating better style transfer than recent, state of the art (SotA) methods, and recognition tests, showing our results remain easier to recognize than those produced by others. Finally, we report a user study, testifying to the improved photo-realism of our results compared to existing methods. 

\begin{figure*}
    \centering
     \includegraphics[clip,trim=0mm 0mm 0mm 0mm,width=\linewidth]{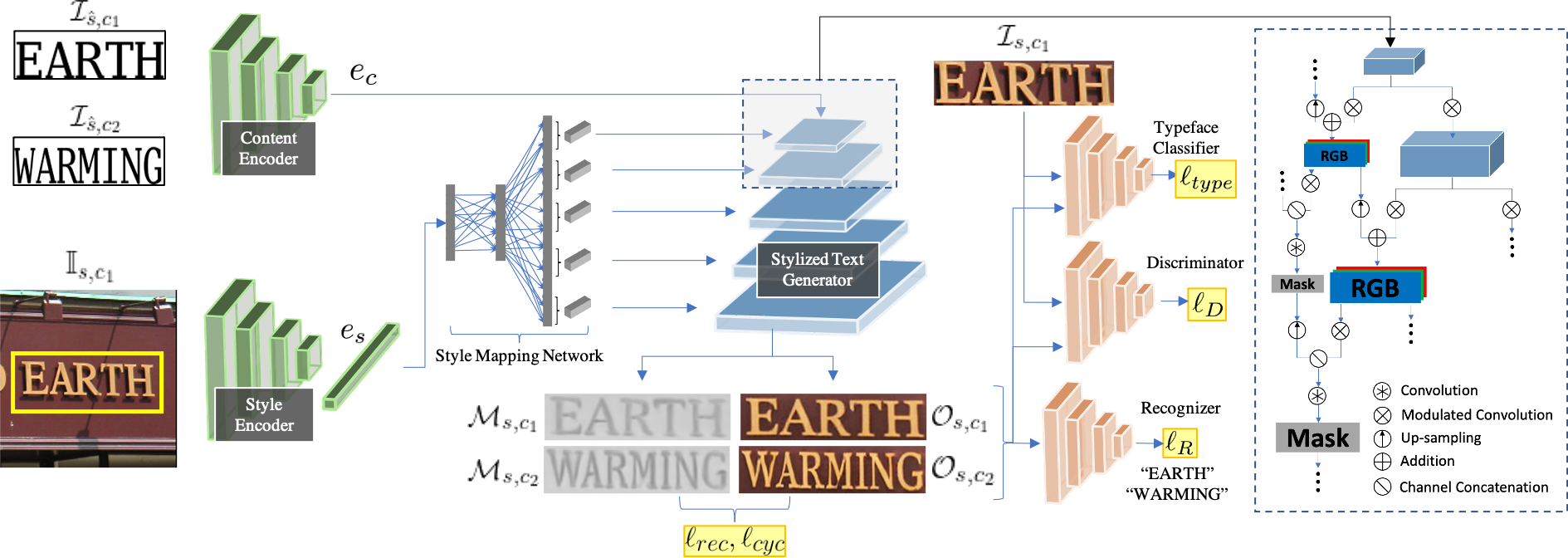}
    \caption{{\bf Overview of our proposed approach, at training.} Content and style encoders, in green, are detailed in Sec.~\ref{subsec:encoders}. The mapping network and generator are explained in Sec.~\ref{sec:stg}. Our loss functions, in yellow, are explained in Sec.~\ref{sec:train}.}
    \label{fig:figure_2_arch}
\end{figure*}

\section{Related work}
\label{sec:relatedworks}
\noindent{\bf Generative image modelling and style transfer.} Models trained to generate photo-realistic content are rapidly improving, based on the progress made on adversarial networks (GAN)~\cite{Goodfellow14} and variational autoencoders (VAE)~\cite{Kingma14}. Isola et al.~\cite{Isola17} proposed the {\em pix2pix} method which uses conditional generative adversarial networks (cGAN) to control the generated content. Zhu et al.~\cite{Zhu17} extended this approach to unpaired, image-to-image translation by learning a mapping between two domains based on cycle consistency. In our work we are interested in performing style transfer in an image-to-image translation setting. Style transfer~\cite{GatysEB16, Johnson16, Huang17} methods extract style from one image and apply it to new content. Huang et al.~\cite{Huang17} described adaptive instance normalization (AdaIN) layers for better infusion of style into intermediate feature maps of the generation process. The idea of injecting learned styles using AdaIN layers has motivated many methods in unsupervised (MUNIT)~\cite{HuangLBK18}, few-shot (FUNIT)~\cite{liu2019few}, multi-domain~\cite{choi2020stargan} setting in natural scene objects. These methods process object classes that 
have an abundance of intra-class examples from the same
``style'' (e.g., cats, dogs). These settings were leveraged to design their multi-task discriminator, style code learning. By comparison, we have far less examples of individual styles, cannot depend on few-shot learning setting in real world scene text images, and the definition of styles could be much more arbitrary. We thus rely on self-supervision via our novel loss functions which preserves desired textual content and style. Importantly, FUNIT was shown~\cite{kang2020ganwriting} to fail in translating handwriting styles.

Karras et al.~\cite{Karras18} generated high-res and high quality content by progressively growing networks in training. Subsequent work introduced StyleGAN, which disentangles an input latent vector in an intermediate latent space, allowing control of high-level attributes and stochastic variations~\cite{KarrasLA19}. This work was further extended by StyleGAN2~\cite{KarrasLAHLA20} which improved generator normalization and proposed path length regularization for better training. We build on the StyleGAN2 architecture, proposing multiple novelties to adapt it to text images.

\minisection{Text generation and style transfer} These methods attempt to capture text styles from source images and apply them to new content. Yang et al.~\cite{YangLLG17} represented styles using textures from image patches of individual glyphs. Their method, however, is computationally expensive. MC-GAN~\cite{AzadiFKWSD18} reduced computation by using stacked cGANs with two sub-networks, capturing text content and style for individual glyphs. Yang et al.~\cite{Yang0WG19} disentangled glyph style and content using autoencoders. Another recent work in this domain is FET-GAN~\cite{li2020fet} which proposed a K-Shot learning for learning styles and uses AdaIN layers for style injection. These methods are limited to individual glyphs and unsuitable for learning style information {\em in the wild}, as we propose. Moreover, FET-GAN uses multiple target images for supervision; we use a single text box and no external style supervision / labels. Although they present results in an unsupervised too, the quality is quite limited. Without style supervision, we, by contrast, depend on indirect signals, captured by our novel losses in an self-supervised manner.

Spatial fusion GAN (SF-GAN)~\cite{ZhanZL19} generates text images by superimposing a foreground content image, transformed to match the style and geometry of a background image. This approach is aimed more at pure text synthesis on image regions without text, whereas we aim at style transfer, replacing existing content with new, using the same style. Gomez et al.~\cite{GomezBGGKR19} used neural style transfer to selectively stylize textual regions of a content image, rendered in a particular style, based on a style image. Their method, however, cannot generate new content, only modify the style of existing text. We formulate the problem more generally, as one-shot disentanglement of style from content and style transfer to new content.
 

\minisection{Scene text editing} Scene text editors typically transfer the text style of an input image to target content, transfer the background style to the target background, and then blend the two. Recent deep learning--based examples include SRNet~\cite{WuZLHLDB19}, which offers an end-to-end trainable style retention network. SWAPText~\cite{YangHL20}, extended SRNet by proposing to additionally model the geometric transformation of the text by using $N$ fiducial points. These methods require target style images for training, thereby being restricted to training on synthetic data. Our method is self-supervised and so does not require target style labels for training and can easily train on real photos. 

Finally, STEFANN~\cite{RoyBG020} recently proposed a font adaptive neural network that replaces individual characters in the source image using a target alphabet. This approach assumes per-character segmentation which is impractical in many real world images. 

\minisection{Handwritten text generation} Handwritten text is challenging due to the high inter-class variability of text styles -- from writer to writer -- and intra-class variability of same writer styles. Dense text CycleGAN~\cite{ChangZPM18} translates handwritten Chinese characters from a machine font to particular handwritten styles. Alonso et al.~\cite{AlonsoMM19} proposed an offline handwriting generator for fixed-size word images. ScrabbleGAN~\cite{FogelACML20} used a fully-convolutional handwritten text generation model that can handle arbitrary text lengths. Most of these methods generate text from random style vectors and so are not directly comparable with our work. Kang et al.~\cite{kang2020ganwriting} used three learning objectives to model handwritten text style, content, and appearance. Their work formulates the training of styles in a K-shot manner and also uses a supervised writer classification network to learn the target style of an author. In contrast, our work only uses one single style image (both training and testing), and we also do not rely on external supervision for writer styles.

One of the closest work to ours is the recent work from Davis et al.~\cite{davis2020text} which also uses a StyleGAN-based generator for generating handwritten lines with a self-supervised training criterion and utilizes perceptual loss, and an OCR-based recognition loss function. The method is more relevant for the handwriting domain since the notion of spaced text using CTC output is applicable for horizontally-aligned text and would not work well for scene text where rotation and curved texts are quite common. In contrast, our method is seamlessly applicable for both the scene text and handwritten domains. We also use a different form of content encoding, train the perceptual classifier (referred as typeface classifier in our work) differently and generate a foreground mask in a self-supervised manner to better learn the nuances of different textual styles. In Sec.~\ref{subsec:expSceneText} \&~\ref{subsec:userStudy}, we show that TSB surpasses this work both quantitatively and in an user study for the task of generating handwritten word images.



\section{Methodology}
\label{sec:method}
Fig.~\ref{fig:figure_2_arch}, illustrates the proposed TSB architecture. We formulate training in a self-supervised manner where we do not have target style supervision and only use the original style image; our framework is designed to supervise itself in seeking photo-realistic results. We do, however, assume to have the ground truth content of each word box when training (the text appearing in it). During inference, we take a single source style image and new content (i.e., a character string), and generate a new image in the source style with the target content.  


We denote the input {\em style image} along with its context by $\mathbb{I}_{s,c} \in \mathbb{R}^{H\times W \times 3}$ and its implicit {\em style} as $s$. The {\em content} string is denoted by $c$. As we explain in Sec.~\ref{subsec:encoders}, we represent $c$ using a synthetically rendered image, $\mathcal{I}_{\hat{s},c}$. 

Our entire framework consists of seven networks. We use a {\em style encoder} $(F_s)$ and {\em content encoder}, $(F_c)$, to convert the input style and content images to latent representations $e_s$ and $e_c$ respectively (Sec.~\ref{subsec:encoders}). Given $e_s$, our {\em style mapping network}, $M$, is used to produce multi-scale style representations, $w_{s,i}$, which are then processed by a {\em stylized text generator} network, $G$ (Sec.~\ref{sec:stg}). Finally, our framework is trained using multiple losses which involve a {\em style loss} computed using a {\em typeface classifier} pre-trained on a limited set of synthetic fonts, $C$ (Sec.~\ref{sec:loss:perecptual}), a {\em content loss}, which uses a pre-trained OCR recognition model, $R$ (Sec.~\ref{sec:loss:content}), and a {\em discriminator}, $D$ (Sec.~\ref{sec:loss:combined}). 

\subsection{Style and content encoders}
\label{subsec:encoders}
We extract latent style and content representations given a text image and an input content string. To this end, we use two {\em encoders}: for style, $(F_s)$, and content, $(F_c)$. Thus, style representation is produced by $e_s=F_s(\mathbb{I}_{s,c})$ and content representation by  $e_c=F_c(\mathcal{I}_{\hat{s},c})$. 

Our content is a character string, e.g., a word (``EARTH''). To simplify training of the two encoders, we opt for a unified input representation which we use for both $(F_s)$, and content, $(F_c)$: a word image. Given a character string $c$ for content, we therefore render it using a standard font for the text, Verily Serif Mono, on a plain white background, producing the image $\mathcal{I}_{\hat{s},c}\in \mathbb{R}^{64\times W}$, where $\hat{s}$ denotes the standard font (Fig.~\ref{fig:figure_2_arch}). Both encoders use a adapted version of ResNet34 architecture~\cite{HeZRS16} with modifications in the last layers. The content representation $e_c \in \mathbb{R}^{512 \times 4 \times W}$ is computed prior to the average pooling layer so that one can preserve the spatial properties. Here, $512$ is the number of channels in last convolutional feature block. 

The style encoder is given a localized scene image ($\mathbb{I}_{s,c}$) along with a word bounding box, denoting the word image from which the style is learned. We found the use of localized scene image (word image + context) helpful since it enables preserving the aspect ratio of input word image and also brings additional contextual features for learning background style components. The penultimate layer of style encoder uses a region of interest (RoI) align operator as described in Mask R-CNN~\cite{he2017mask} to pool features coming from the desired word region, thereby reducing the style tensor to a fixed, 512D representation. Note that, both these encoders are fully convolutional and can handle variable sized inputs and outputs.


\subsection{Generator and style mapping network}
\label{sec:stg}
For our stylized text generator, $G$, we use a model based on StyleGAN2~\cite{KarrasLAHLA20}, due to the exquisite, photo-realistic images it generates in domains such as faces and outdoor scenes. For our goal of generating photo-realistic text images, however, the design of StyleGAN2 has two important limitations. 

First, StyleGAN2 is an unconditional model which generates images by sampling a random latent vector, $z\in \mathcal{Z}$. We, however, need to control the output based on two separate sources: Our desired new text content and the existing style. A second limitation relates to the unique nature of stylized text images. Faithfully representing text styles involves a combination of global information -- e.g., the spatial transformation of the text, its color palette -- with detailed, fine-scale information such as the existence of serifs~\cite{samara2004typography} or minute variations of individual penmanship styles~\cite{hood1992pen}.   

We address these limitations jointly, by introducing key novelties to the standard StyleGAN2 model, as illustrated in Fig.~\ref{fig:figure_2_arch}. We condition our generator on our content, $e_c$, and style, $e_s$, representations. The content representation is directly fed as the input to the first layer of the generator instead of providing the learned tensor which was used in the original StyleGAN2 model. 

We further handle the multi-scale nature of text styles by extracting {\em layer-specific style information}. To this end, we introduce a {\em style mapping network}, $M$, which converts $e_s$ to layer-specific style representation, $w_{s,i}$, where $i$ indexes the layers of the generator which are then fed as the AdaIN normalization coefficient~\cite{Huang17} to each layer of the generator. Thus, we allow the generator to control both low and high resolution details of the text appearance to match a desired input style. 

To further address the issue of preserving the target style and its minute variations, we additionally model the generation of foreground components of the image (pixels belongs to the text). As shown in Fig.~\ref{fig:figure_2_arch}, we generate a {\em soft} mask ($\mathcal{M}_{s,c}$) along with the generation of target image. The zoomed-out figure shown in the right part of architecture illustrates the novelties introduced to the generator layers to compute the mask. The mask is generated at each layer by taking channel wise concatenated inputs from the current layer RGB image and the previous layer mask. It uses the sigmoid activation function to compute soft weights for each pixel denoting its probability to foreground (text region) or background region. The final mask provides a soft semantic segmentation of the generated image and is later used to construct the loss functions. Note that the mask is also learned in a self-supervised manner, in particular, no mask labels are provided during training. The advantage of learning mask is validated in the ablation study conducted in Sec~\ref{subsec:expSceneText}. Finally, consequent to these changes, we eliminate the use of the noise vector input of the standard StyleGAN2.

\section{TSB Training}
\label{sec:train}
We do not assume that we have style labels for training our method on real photos. In fact, because of the essentially limitless variability of text styles, it is not clear what high level parameters can be used to capture these styles. We therefore take an indirect approach, decoupling our loss into components which, taken together, allow for effective self-supervision of our training.

As shown in the Fig.~\ref{fig:figure_2_arch}, we provide two content inputs ($\mathcal{I}_{\hat{s},c_1},\mathcal{I}_{\hat{s},c_2}$) while training the network along with the input scene image $\mathbb{I}_{s,c_1}$ which contains the stylized word. We assume the ground truth text content for the input $c_1$ known while training. The generator produces the target images ($\mathcal{O}_{s,c_1},\mathcal{O}_{s,c_2}$) along with their masks ($\mathcal{M}_{s,c_1},\mathcal{M}_{s,c_2}$). Note that, due to self-supervision, the ground truth for $\mathcal{O}_{s,c_1}$ is $\mathcal{I}_{s,c_1}$ (cropped word image from $\mathbb{I}_{s,c_1}$), however the ground truth for $\mathcal{O}_{s,c_2}$ remains unknown. We use the following loss functions, explained next, to enable learning text styles along with their masks, {\em indirectly}, in a self-supervised manner.

\subsection{Text perceptual losses} \label{sec:loss:perecptual}
We measure how well our generator captures the style of input text by using an approach analogous to perceptual loss~\cite{GatysEB16,HuangLBK18,WuZLHLDB19, YangHL20}. Specifically, we assume a pre-trained typeface classification network, $C$. Importantly, we make {\em no assumption} that the fonts $C$ was trained to classify would appear in the photos we use to train and evaluate our TSB. Network $C$ is only used to provide a perceptual signal for training. In particular, this network is trained to identify standard synthetic fonts.

For details of the synthetic data used to train $C$, see Sec.~\ref{sec:dataset}. We use this data to train a VGG19 network~\cite{SimonyanZ14a} to produce a font class given a word image. Training used the softmax loss with one-hot encoded font class labels. Note that the synthetic data is only used to pre-train $C$ while all other components of the TSB architecture are trained using real-world datasets where there is no availability of target font information.

Given a word image, we extract the following values from $C$ to compute our {\em text-specific perceptual loss}, $\ell_{type}$: 
\begin{gather}
\ell_{type} = \lambda_{1}\ell_{per} + \lambda_{2}\ell_{tex} + \lambda_{3}\ell_{emb},\\
\ell_{per}=\mathbb{E}[\sum_i\frac{1}{M_i}\parallel \phi_i(\mathcal{I}_{s,c_1})-\phi_i(\mathcal{O}_{s,c_1})\parallel_1],\\
\ell_{tex}=\mathbb{E}_i[\parallel G_i^\phi(\mathcal{I}_{s,c_1}) -G_i^\phi(\mathcal{O}_{s,c_1})\parallel_1],\\
\ell_{emb}=\mathbb{E}[\parallel \psi(\mathcal{I}_{s,c_1})-\psi(\mathcal{O}_{s,c_1})\parallel_1].
\end{gather}
Here, $\ell_{per}$ corresponds to the perceptual loss computed from the feature maps at layer $i$ denoted as $\phi_i$ and $M_i$ is the number of elements in the particular feature map which is used as normalization. Next, $\ell_{tex}$ is a texture loss (also known as style loss~\cite{GatysEB16}), computed from the Gram matrices $G_i^{\phi}=\phi_i\phi_i^T$ of the feature maps. Both these losses are computed on the initial set of layers of VGG which are: $\text{relu1\_1, relu2\_1, relu3\_1, relu4\_1, relu5\_1}$.

We additionally use an embedding-based loss which we compute from feature maps, $\psi$, of the penultimate layer of this trained typeface classification network. We use the perceptual and texture losses to learn more about the background style information while the embedding loss provides font-level cues for the generated image. Note that the perceptual losses are only computed for the output image corresponding to original content $c_1$.


\subsection{Text content loss}\label{sec:loss:content}
We use a pre-trained text recognition network, $R$, to evaluate the content of the generated images. We use the output string estimated by $R$ to compute a loss, $\ell_{R}$, which reflects how well the generator captured the desired content string, $c_1,c_2$ simultaneously on both target images ($\mathcal{O}_{s,c_1},\mathcal{O}_{s,c_2}$) and their masks ($\mathcal{M}_{s,c_1},\mathcal{M}_{s,c_2}$). Ideally a recognizer should only focus on the text content (foreground element) of the image irrespective of the its background elements. Therefore, constraining it to recognize the same content string on both the target generation and its mask allows us to align it well. In practice, we use an existing word pre-trained recognition model by Baek et al.~\cite{BaekKLPHYOL19}. Of the ones proposed in that paper, we chose the model with the following configuration, though we did not optimize for this choice: (1) spatial transformation network (STN) using thin-plate spline (TPS) transformation, (2) feature extraction using ResNet network, (3) sequence modelling using BiLSTM, and (4) an attention-based sequence prediction. This OCR was favored above other methods which may be more accurate~\cite{liao2020mask}, due to its simplicity and ease of integration as part of our approach. 

The content loss is computed by measuring the cross entropy between the sequence of characters in the input string, $c_1,c_2$, the predicted string, $c'_1,c'_2$ respectively and are represented as one-hot vectors.

\subsection{Reconstruction losses} 
\label{sec:loss:reconstruct}
A reconstruction loss function is one of the important loss criteria in self-supervised learning. In this work, we capitalize on the learned mask image to disentangle the foreground and background pixels of the generated image, thereby allowing us to apply reconstruction losses effectively. We use two reconstruction losses: $\ell_{rec}$ and $\ell_{cyc}$. Here, $\ell_{rec}$ represents the differences between the output-generated image in style $s$ and content $c_1$, $\mathcal{O}_{s,c_1}$, and the cropped input-style example, $\mathcal{I}_{s,c_1}$. 

To further enforce the presence of target style in generation we perform a cyclic reconstruction on the generated image. We compute a {\em fake style} vector, $e_s'=F_s(\mathbb{O}_{s,c_1})$, and generate $\mathcal{O}_{s',c_1}$. Here, $\mathbb{O}_{s,c_1}$ is a localized stitched image along with the context region taken from the input $\mathbb{I}_{s,c_1}$. The loss, $\ell_{cyc}$, is computed between $\mathcal{O}_{s',c}$ and $\mathcal{I}_{s,c}$. We use an $L_1$ loss criterion for both reconstruction losses and split these functions on the foreground and background region separately using the generated masks. This allows the network to learn fine-grained variations of the style present in the foreground region.

\subsection{The combined loss}\label{sec:loss:combined}
Our full loss is given by the following expression:
\begin{equation}
    \ell = \ell_{D} + \lambda_{4}\ell_{R} +  \ell_{type} + \lambda_{5}\ell_{rec} + \lambda_{6}\ell_{cyc}.
    \label{eq:loss}
\end{equation}
Here, $\ell_{D}$ denotes the discriminator-based adversarial loss. We use the non-saturating loss function~\cite{Goodfellow14} with $R_1$ regularization for the discriminator and path length regularization~\cite{KarrasLAHLA20} for the generator. Other terms appear in previous sections: $\ell_{R}$ is the content loss of Sec.~\ref{sec:loss:content}, $\ell_{type}$ refers to losses computed by the typeface classification network, $C$, Sec.~\ref{sec:loss:perecptual}. Finally, $\ell_{rec}$ and $\ell_{cyc}$ are the reconstruction and cyclic reconstruction losses, respectively (Sec.~\ref{sec:loss:reconstruct}). While training our network, we keep the weights of the pre-trained networks, $R$ and $C$, frozen. We empirically set the values of the balance factors $\lambda_1-\lambda_6$ and report in Sec.~\ref{subsec:implDetails}.

\subsection{Inference}\label{sec:inference}
During inference, we present a novel localized style word image (or a cropped word) $\mathbb{I}_{s,c}\in\mathbb{R}^{H \times W \times 3}$ as input to the style encoder and the target content as a synthetically rendered image $\mathcal{I}_{\hat{s},c} \in \mathbb{R}^{64 \times W}$ to the content encoder. The learned representation $e_s,e_c$ from these encoders are given as the conditional data to the generator which generates the target image, $\mathcal{O}_{s,c}\in \mathbb{R}^{64 \times W \times 3}$ and its mask $\mathcal{M}_{s,c}\in \mathbb{R}^{64 \times W}$. Please note that all the network components in the inference pipeline are fully convolutional and therefore support variable length generation.

\section{Imgur5k Handwriting set}\label{sec:imgur}
Although there are many data sets for training and evaluating OCR systems on printed and scene text (see Singh et al. for a recent survey~\cite{textocr}), we found that existing collections of handwritten word images to be limited in their variability (See Table~\ref{tab:imgurdataset}). To facilitate training and testing of our TSB, we, therefore, collected and annotated $\sim$135K handwritten English words from 5K images originally hosted publicly on Imgur.com. Images were selected based on their affiliation with online handwriting communities, which favor unique scripts and upload snapshots of novel handwriting. Collected images were annotated for word-level bounding boxes and the strings contained within. Each image was assigned to multiple (up to five) annotators and the annotation average of word bounding boxes and highest agreement of labeled content strings was used to eliminate spurious data. We use a split containing $\sim$108K word images for training, $\sim$13K for validation, and $\sim$14K images for testing TSB network; all splits are document-exclusive.

Imgur5K is unique among handwriting datasets in that it meets all of the following criteria: its content is structured into words and sentences instead of isolated characters, it is not restricted to a single source domain, it has an author:document ratio of approximately 1, and it contains a vast array of image formats, resolutions, framings, and backgrounds. 

\begin{figure}[t]
\begin{center}
\includegraphics[width=\linewidth]{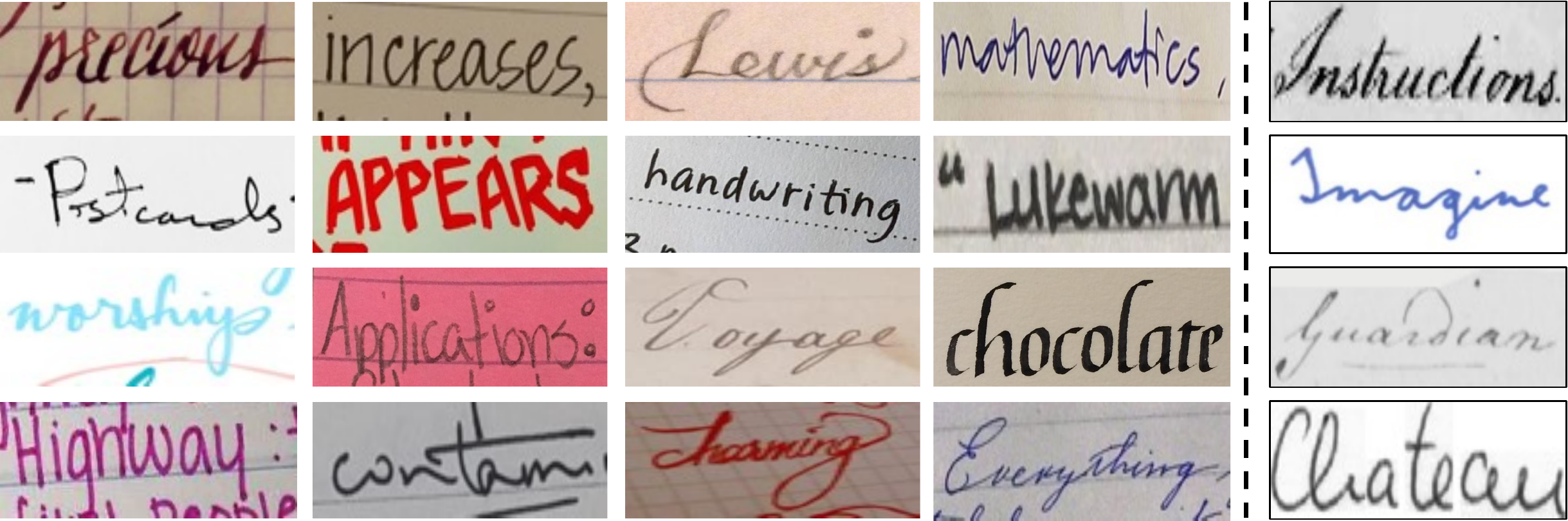}
\end{center}
   \caption{{\bf Imgur5K word images}. (Cols 1-4) Sample word images from the Imgur5K dataset. (Col 5) A single sample word image from each of the datasets (top to bottom) GW, CVL, Bentham, IAM respectively.}
\label{fig:imgur5k}
\end{figure}

\begin{table}[t]
    \centering
    \resizebox{\columnwidth}{!}{%
    \begin{tabular}{lccrcclc}\toprule
         \textbf{Dataset} & \textbf{\#Words}  &  \textbf{\#Lexicon} & \textbf{\#Writers}  & \textbf{\#Pen Styles} & \textbf{Bg. Clutter} \\\midrule
         GW~\cite{rath2007word} & 4,894 & 1.4K & $1^{*}$ & Low & Low \\
         CVL~\cite{kleber2013cvl} & 101,069 & 292 & $\sim 311$ & Low & Low \\
         Bentham R0~\cite{sanchez2014icfhr2014} & 110,000 & 9.5K &$1^{*}$ & Low & Medium \\
         IAM~\cite{MartiB02} & 115,320  & 13.5K &  $\sim 657$  & Medium & Low \\ \hline
         Our Imgur5K & \textbf{135,375} & \textbf{27K} &  $\sim \textbf{5305}$ & \textbf{High} & \textbf{High} \\
         
         \bottomrule
    \end{tabular}
    }
    \caption{{\bf Comparison of popular datasets of handwritten English words}. $1^*$ denotes documents written primarily by one author (though an unknown number of assistants may have also contributed to the set.)}
    \label{tab:imgurdataset}
\end{table}

Fig.~\ref{fig:imgur5k} shows a sample set of word images from this collection. Table~\ref{tab:imgurdataset} compares Imgur5K to other popular handwritten datasets in English, under different attributes. Clearly, Imgur5K has much more writer variability and possesses real world challenges in terms background clutter, variability in pen types and style ranges from noisy
writing to precise calligraphy written by experts. The last column of Fig.~\ref{fig:imgur5k} shows a typical sample taken from different existing datasets in order to highlight the in-wild nature of Imgur5K. Other datasets are mainly captured either in a constrained setting or written by one author alone. Finally, Imgur5K is already available online from: \footnote{\url{https://github.com/facebookresearch/IMGUR5K-Handwriting-Dataset}}.


\section{Experiments}
\label{sec:exp}
We rigorously tested our proposed TSB architecture and report qualitative and quantitative results, as well as a user study, comparing our results with previous work. 

\subsection{Implementation details}
\label{subsec:implDetails}
We use a fixed-size localized style image of size $256\times 256$. The localized image is cropped out from the larger scene image along with its context while preserving the original aspect ratio of the word present inside it. We also assume the ground-truth rectangular word bounding box information available. The synthetically rendered content image is of dimension $64 \times 256$ for training, and during inference it uses a variable width image of dimension $64 \times W$. The style and content encoders, $(F_s)$ and $(F_c)$, respectively, (Sec.~\ref{subsec:encoders}), use ResNet34~\cite{HeZRS16}, following the format proposed by Zhanzhan et al.~\cite{ChengBXZPZ17} for deep text features and the modification as explained in Sec.~\ref{subsec:encoders}.


We use StyleGAN2~\cite{KarrasLAHLA20} for our generator (Sec.~\ref{sec:stg}). We base our generator on the StyleGAN2 variant with skip connections, a residual discriminator, and without progressive growing. We adapt this architecture to also produce foreground masks for the generated image as shown in the Fig.~\ref{fig:figure_2_arch}. We also modified the input dimensions to generate output images of size $64\times 256$. The learned content representation $e_c \in \mathbb{R}^{4\times 16}$ is given as input to the first layer of the generator. As explained in Sec.~\ref{sec:stg}, we do not use noise inputs, instead conditioning the output on our latent style and content  vectors.


Our models were all trained end-to-end, with the exclusion of the pre-trained networks -- the typeface classifier, $C$, of Sec.~\ref{sec:loss:perecptual} and recognizer, $R$ of Sec.~\ref{sec:loss:content} -- which were kept frozen. We use the Adam optimizer with a fixed learning rate of 0.002 and batch size of 64. We empirically set the relative weights of the different loss functions as: $\lambda_1=1.0, \lambda_2=500.0, \lambda_3=1.0, \lambda_4=1.0, \lambda_5=10.0, \lambda_6=1.0$. 

Finally, our method is implemented using the PyTorch distributed  framework~\cite{pytorch}. Training was performed on 8GPUS with 16GB of RAM each.

\subsection{Datasets}
\label{sec:dataset}
Our experiments use a variety of datasets, representing real and synthetic photos, scene text and handwriting. Below we list all the datasets used in this work. The sets and annotations collected as part of this work shall be publicly released, as well as any test splits used in our experiments. 

\minisection{Synthetic data} We use three different synthetic datasets in this work. 

\noindent \textit{SynthText}~\cite{GuptaVZ16}: We use the SynthText in the wild dataset for training our architecture on synthetic data in a self-supervised manner. The synthetically-trained model is later used for purpose of performing an ablation study (Sec.~\ref{subsec:expSceneText}) on our architecture. 
\vspace{0.1cm}

\noindent \textit{Synth-Paired dataset}: This is the test synthetic dataset prepared for the ablation study (Sec.~\ref{subsec:expSceneText})  using the pipeline presented by SRNet~\cite{WuZLHLDB19} where we can produce target (new content) style word images following the source style. Note that, the above two datasets and the synthetically TSB trained model is only used for the ablation study. All other experiments use real world datasets where the supervision of target style is unavailable.

\vspace{0.1cm}
\noindent \textit{Synth-Font dataset}: This is a separate word level synthetic set of around 250K word images, sampled from $\sim$2K different typefaces using again the pipeline presented by SRNet~\cite{WuZLHLDB19}. This separate set was used in pre-training our typeface classification network, $C$, as mentioned in Sec.~\ref{sec:loss:perecptual}. 


In addition to the synthetic sets, we used collections of real images. These sets are described below, with the exception of our own, Imgur5K, detailed in Sec.~\ref{sec:imgur}.

\minisection{ICDAR 2013}~\cite{KaratzasSUIBMMMAH13} This set is part of the ICDAR 2013 Robust Reading Competition. Compared to other real datasets, ICDAR 2013 images are of higher resolution with prominent text. There are 848 and 1,095 word images in the original train and test sets. 


\minisection{ICDAR 2015}~\cite{KaratzasGNGBIMN15} Released as part of the ICDAR 2015 Robust Reading competition, this set was designed to be more challenging than ICDAR 2013. Most of the images in this set have low resolution and viewpoint irregularities (e.g., perspective distortions, curved text).

\minisection{TextVQA}~\cite{singh2019towards} The dataset was collected for the task of visual question answering (VQA) in images. It contains 28,408 scene photos sourced from the OpenImages set~\cite{OpenImages2}, with 21,953 images for training, 3,166 for validation, and 3,289 for testing. While the purpose of the dataset is VQA, it contains a large variety of scene text in different, challenging styles, more so than other sets. We annotated this set with word polygons and recognition labels. 

\minisection{IAM Handwriting Database}~\cite{MartiB02} This set contains 1,539 handwritten forms, written by 657 authors. IAM offers sentence-level labels, lines, and words. In our work, we only use word-level annotations. We use the official partition for writer independent text line recognition which splits forms into writer-exclusive training, validation, and testing sets.



\subsection{Evaluation measures}
\label{subsec:evalMeas}
We follow the same quantitative evaluation measures as previous scene text editing methods~\cite{WuZLHLDB19, YangHL20}. We compare target style images with generated results using these measures: (1) {\em Mean square error} (MSE), the $l_2$ distance between two images; (2) {\em structural similarity index measure} (SSIM)~\cite{WangBSS04}; (3) {\em Peak signal-to-noise ratio} (PSNR). Low MSE scores and high SSIM and PSNR scores are best. 

These metrics can only be used with a synthetic image set, where we can generate a corresponding target style image. To test on real photos, where we do not have a prediction of how new text would look with an example style, we measure text recognition accuracy. It is computed as 
\begin{equation}
Acc = \frac{1}{\#test}(\sum_i \mathbb{I}(R(\mathcal{O}_{s_i,c_i})==c_i)), \label{eq:recadd}
\end{equation}
or, the number of times a predicted string is identical to the actual string $c_i$. To this end, we use our pre-trained text recognition network, $R$, (Sec.~\ref{sec:loss:content}). 

To evaluate real handwritten images, we use methods from the GAN literature, also adopted by previous handwritten text generation methods~\cite{AlonsoMM19,FogelACML20, davis2020text}: (1) $\text{Fr\`{e}chet Inception Distance}$ (FID)~\cite{HeuselRUNH17} calculates the distance between the feature vectors of real and generated images, and, (2), the $\text{Geometric Score}$ (GS)~\cite{KhrulkovO18}, which compares geometrical properties between the real and generated manifolds. Similar to others~\cite{davis2020text}, we evaluate GS on a reduced set, due to its computational costs, randomly sampling 5K images in both real and generated sets. We resized the images to a fixed size ($64\times 256$) for GS computation. FID was computed on sampled real and generated images the size of the test sets using variable-width images ($299\times W$) following the protocol presented in~\cite{davis2020text}. Lower scores on both metrics are better. Finally, we report results of a user study which compares the visual quality of our results and those of previous works. Details of this study are discussed in Sec.~\ref{subsec:userStudy}. 


\begin{table}[!t]
    \centering
    \resizebox{\columnwidth}{!}{%
    \begin{tabular}{lcccccc}\toprule
        \textbf{Loss/Method}  & Scale & Mask & {\sc \textbf{mse}} $\downarrow$ & {\sc \textbf{ssim}} $\uparrow$ & {\sc \textbf{psnr}} $\uparrow$ & {\sc \textbf{fid}} $\downarrow$ \\\midrule
        $\ell_D$   & M & \ding{51} & 0.0857 & 0.2201 & 12.59 & 152.03\\
        $\ell_D + \ell_R$  & M & \ding{51} & 0.0672 & 0.1754 & 13.015 & 165.41 \\
        $\ell_D + \ell_R + \ell_{rec}$  & M & \ding{51} & 0.0180 & 0.4164 & 18.68 & 86.79 \\ 
        $\ell_D + \ell_R + \ell_{rec}  + \ell_{cyc}$ & M & \ding{51} & \textbf{0.0162} &  \textbf{0.4434} & \textbf{19.23} & 99.45\\ \midrule
        $\ell_D + \ell_R + \ell_{rec}  + \ell_{cyc} + \ell_{type}$ & S & \ding{51}   & 0.0199 & 0.4041 & 18.23 & 87.68 \\
        $\ell_D + \ell_R + \ell_{rec}  + \ell_{cyc} + \ell_{type}$ & M & \ding{55}   & 0.0172 & 0.4416 & 19.07 & 90.67 \\
        $\ell_D + \ell_R + \ell_{rec}  + \ell_{cyc} + \ell_{type}$ & M & \ding{51} & 0.0196 & 0.4174 & 18.35 & \textbf{79.49} \\
        \bottomrule
         
    \end{tabular}
    }
    \caption{{\bf Ablation study} comparing the influence of different loss functions on our TSB results.}
    \label{tab:ablStudy}
\end{table}



\begin{figure*}[t]
\begin{center}
\includegraphics[width=\linewidth]{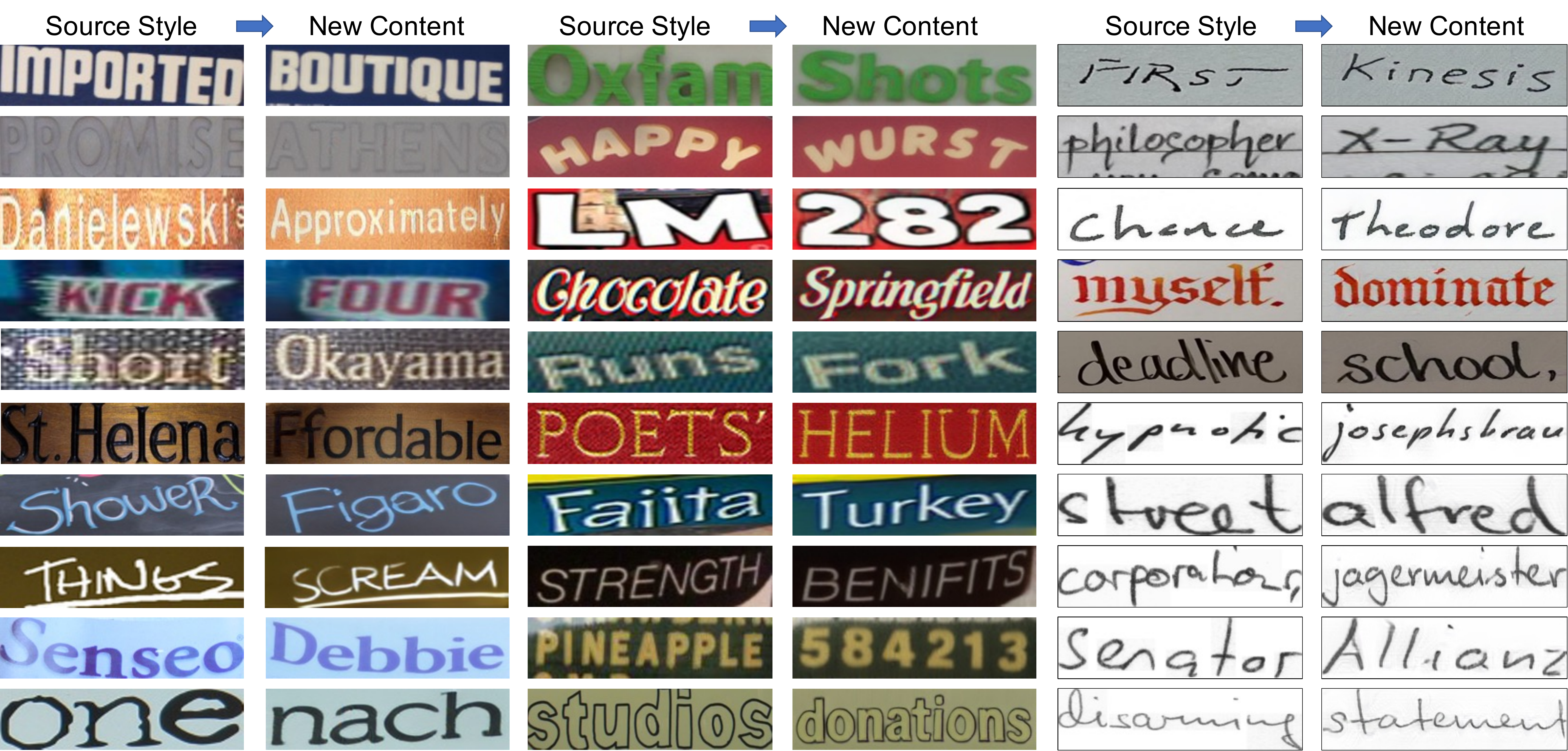}
\end{center}
   \caption{{\bf Word-level style transfer results}. Each image pair shows input source style on the left and output with novel content (string) on the right. All examples are real photos (no synthetic data) taken from ICDAR13~\cite{KaratzasSUIBMMMAH13}, TextVQA~\cite{singh2019towards}, IAM handwriting~\cite{MartiB02}, and the Imgur5K set collected for this work. Note that the handwritten results are of variable widths, but resized to fixed dimension in this figure for better visualization. {\em See supplemental for more results.}}
\label{fig:wordGen}
\end{figure*}

\subsection{Text generation results}
\label{subsec:expSceneText}

\minisection{Ablation study} Table~\ref{tab:ablStudy} provides an ablation study evaluating the effects of the different loss functions, scale of style features and the role of having masks while training TSB (Sec.~\ref{sec:train}). The first three quantitative metrics (MSE, SSIM and PSNR) focus on pixels level differences and may not reflect the true perceived visual quality. Hence, we also add the FID metric which is adopted in many GAN based generative models. The first setting ($\ell_D$) mimics the original style GAN training with the difference that the noise inputs are replaced with conditional style and content vectors. Although the images are produced in a realistic manner, it performs worst in terms of metrics since there are no losses which captures the respective content and style. Using $\ell_D+\ell_R$ while training also produces realistic results but the output style is still inconsistent with the input style. Only when we add the reconstruction loss, $\ell_{rec}$, do the input and output styles became consistent. We improve results even more by adding the cyclic reconstruction loss, $\ell_{cyc}$, which contributed to even better consistency between source and target styles. Note that, the above settings used the multi-scale (M) style representation and utilizes the foreground mask while training.

\begin{figure*}[t]
    \centering
    \includegraphics[width=\textwidth]{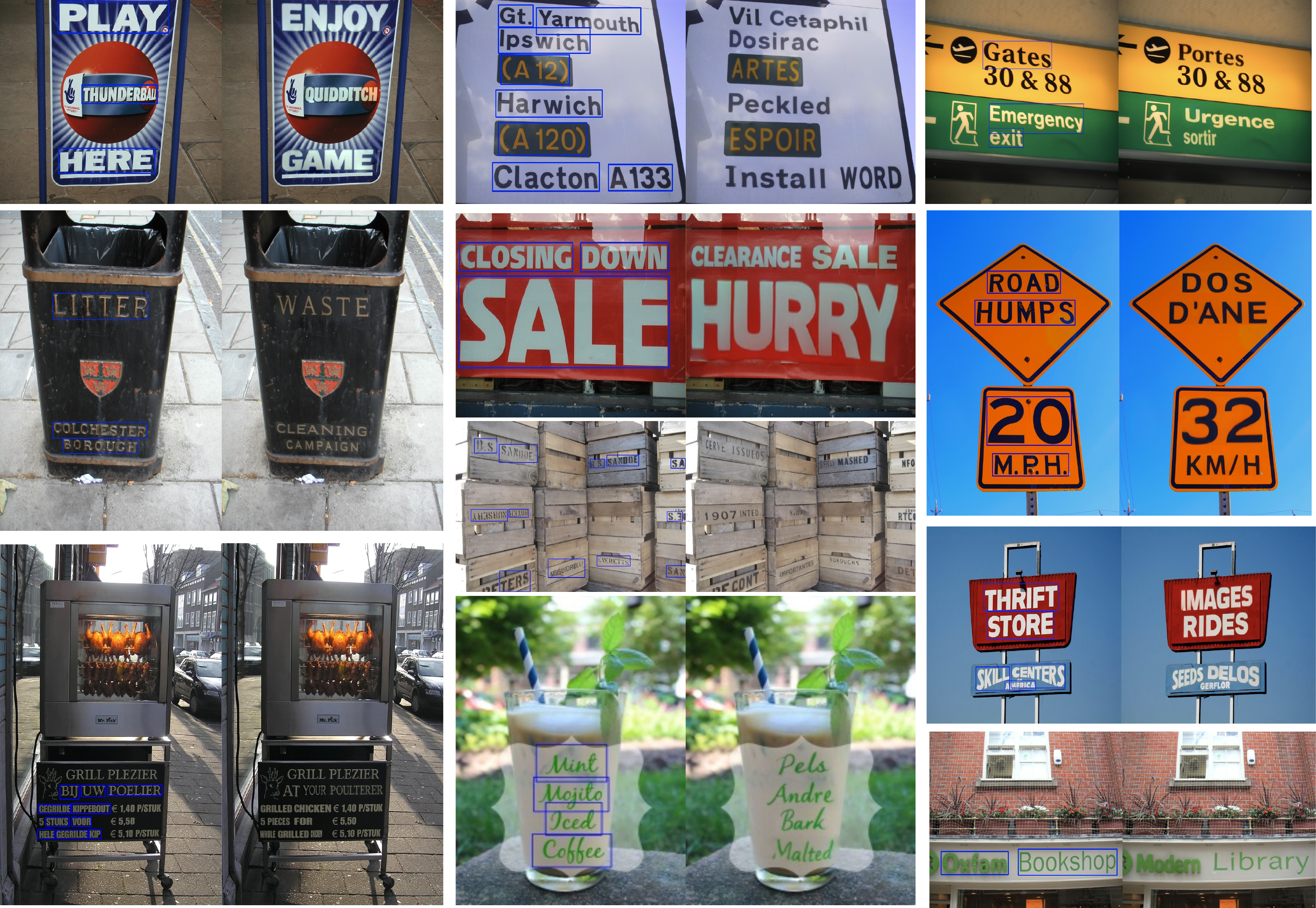}
    \caption{{\bf Scene text editing results.} On left we show the original scene image with word bounding boxes shown in blue rectangles, and on right we present the edited image with text content replaced and blended back using simple Poisson blending~\cite{perez2003poisson} to the scene image. These examples are taken from ICDAR13~\cite{KaratzasSUIBMMMAH13} and TextVQA~\cite{singh2019towards} dataset. {\em See supplemental for more results.}}
    \label{fig:scene-text-edit}
\end{figure*}

The last three rows compares the model trained with the perceptual loss, $\ell_{type}$ in addition to the other loss functions presented before. Here we notice a change in performance where the FID metric starts improving while there is a drop in other metrics such as MSE, SSIM and PSNR. This behaviour is because the latter metrics are biased towards preferring output images which blurry or smooth and penalize the sharp images. We asserted this by performing Gaussian filtering on the output images and noticed the drop in performance across these metrics. Here we consider the FID metric to be more reliable and consistent with human visual inspection. Our best results in terms of FID metric (last row), are produced when we have full text-specific perceptual loss, $\ell_{type}$, of Sec.~\ref{sec:loss:perecptual} with multi-scale style feature and trained using foreground masks. The perceptual and texture losses, $\ell_{per}, \ell_{tex}$, respectively, constrain the activations of the initial layers and the stationary properties of the generated image, making them consistent with input style image. The embedding loss, $\ell_{emb}$, penalises the results based on typeface-level features.

\begin{table}[t]
    \centering
    \resizebox{\columnwidth}{!}{%
    \begin{tabular}{lccc}\toprule
         \textbf{Dataset}  &  \textbf{IC13}~\cite{KaratzasSUIBMMMAH13} & \textbf{IC15}~\cite{KaratzasGNGBIMN15} & \textbf{TextVQA}~\cite{singh2019towards} \\\midrule
         Real~\cite{BaekKLPHYOL19} & 92.3 & 77.6 & 50.3 \\ \hline
         SRNet~\cite{WuZLHLDB19} & 49.8 & 23.4 & 30.3 \\
         SWAPText~\cite{YangHL20}* & 68.3 & 54.9 & - \\ \hline
         Our TSB & \textbf{97.2} & \textbf{97.6} & \textbf{95.0}\\\bottomrule
         
         
    \end{tabular}
    }
    \caption{{\bf Text recognition accuracy} on images from three datasets. {\em Real} is provided as the baseline recognition of $R$ on the real photos. * Denotes results reported by SWAPText~\cite{YangHL20} on an undisclosed subset of ICDAR 13. Since they did not release code, their numbers are not comparable with others and provided here {\em only for reference}.}
    
    \label{tab:textrecog}
\end{table}

\minisection{Recognition accuracy on real photos} Following others, we compare machine recognition accuracy on generated images. Accuracy is measured using Eq.~\eqref{eq:recadd} and we compare with SRNet~\cite{WuZLHLDB19} and SWAPText~\cite{YangHL20}. For these tests, we did not fine-tune our model on the ICDAR 2013 and ICDAR 2015 sets since these provide very few training images. 

Importantly, the test splits used by SWAPText~\cite{YangHL20} and SRNet~\cite{WuZLHLDB19} were not disclosed. Furthermore, SWAPText did not share their code. We consequently provide SWAPText numbers from their paper, only for reference, although they are not directly comparable with ours. We use a third party implementation~\cite{srnetCode} of SRNet along with a pre-trained model trained in a supervised setting for comparing it with our TSB model on the same test images.

Table~\ref{tab:textrecog} reports these recognition results. The first row provides baseline accuracy of our recognizer, $R$ (Sec.~\ref{sec:loss:content}), on the original photos. This accuracy is computed by comparing its output with the human labels available for these sets. Evidently, the recognition engine is far from optimal, yet despite this, serves very well in training our model. Our method clearly generates images with better recognizable text compared with  images generated by SRNet~\cite{WuZLHLDB19} and SWAPText~\cite{YangHL20}. 


\minisection{Qualitative results} Fig.~\ref{fig:wordGen} presents word-level qualitative samples generated by our TSB. We show both the source (input) style box and our generated results with new content. Our method clearly captures the desired source style, from just a single sample. To our knowledge, this is {\em the first time} one-shot text style transfer is demonstrated for both scene text and handwriting domain. 
Fig.~\ref{fig:scene-text-edit}, presents scene text editing results from ICDAR13~\cite{KaratzasSUIBMMMAH13} and TextVQA~\cite{singh2019towards} datasets. The left image is the original scene image along with words marked for replacement (shown in blue bounding boxes) and the right one is edited image using new content by the TSB. We demonstrate these results by selectively stitching back the generated word image (foreground and background separately using masks) back to the source bounding box using Poisson blending. See supplemental for more details.

\begin{table}[t]
    \centering
    \resizebox{\columnwidth}{!}{
    \begin{tabular}{llcc}\toprule
        \textbf{Method} &  \textbf{Dataset} & {\sc \textbf{fid}$\downarrow$} & {\sc \textbf{gs}$\downarrow$} \\\midrule
        Davis et al.~\cite{davis2020text} & IAM words & 104.95 & $1.37 \times 10^{-4}$ \\
        Our TSB  & IAM words & \textbf{44.68} & $\textbf{1.09} \times \textbf{10}^{\textbf{-4}}$  \\\hline
        Our TSB  & Our Imgur5K words & \textbf{47.37} & $\textbf{5.45} \times \textbf{10}^{\textbf{-5}}$  \\
        \bottomrule
    \end{tabular}
    }
    \caption{{\bf Quantitative handwritten results.}}
    \label{tab:hwQuant}
\end{table}

\minisection{Offline handwritten generation results} Table~\ref{tab:hwQuant}, provides quantitative comparisons of generated handwritten texts, comparing our method with Davis et al.~\cite{davis2020text} the recent SotA method designed specifically for generating handwritten text. Since they did not report FID/GS metrics for IAM dataset at word level, we took their official implementation (available from the author's Github repository) along with the pre-trained model for validating these results. We report FID scores where the lower the values better the generation quality. Evidently, here too, our method outperforms the previous work. 


Qualitative examples of handwritten output are provided in Fig.~\ref{fig:wordGen}. We emphasize again, that handwritten styles are learned in a one-shot manner, from the single word example provided for the sourced style while previous methods such as Davis et al.~\cite{davis2020text} uses a much larger image (a sentence or two) to extract the style representation.

\begin{table}[t]
    \centering
    \resizebox{\columnwidth}{!}{%
    \begin{tabular}{llc}\toprule
         \textbf{Domain} & \textbf{Method}  &  \textbf{Original/Generated}\\\midrule
         \multirow{3}{*}{Scene Text} & Real & 87.00\\
         & SRNet~\cite{WuZLHLDB19} & 28.64\\
         & Our TSB & \textbf{58.88}\\\hline
          \multirow{3}{*}{Handwritten} & Real & 87.39\\
          & Davis et al.~\cite{davis2020text} & 37.45\\
         & Our TSB & \textbf{65.76}\\\hline
         \multirow{2}{*}{Comparison} 
         & TSB vs. SRNet~\cite{WuZLHLDB19}  & \textbf{77.70}\\
         & TSB vs. Davis et al.~\cite{davis2020text} & \textbf{77.65}\\
         \bottomrule
    \end{tabular}
    }
    \caption{{\bf User study}. The user study shows clear margin in favor of our results, across scene text and handwriting domains. See Sec.~\ref{subsec:userStudy} for details.}
    \label{tab:userStudy}
\end{table}

\subsection{User study results}
\label{subsec:userStudy}
Text styles and aesthetics are ultimately qualitative concepts, often designed to appeal to human aesthetics. Recognizing that, we also evaluate generated images in a user study, comparing our results with those of SotA methods. Specifically, we tested the quality of scene text generation methods by randomly sampling 60 word images from the ICDAR 2013 set, and used them as source styles for comparing our method with SRNet~\cite{WuZLHLDB19}. We compared the quality of generated handwritten word images by sampling 60 images from different authors present in the IAM dataset. These were used as style source images when comparing our results with those produced by Davis et al.~\cite{davis2020text}. These style images were line instances for Davis et al.~\cite{davis2020text} since the original method showed results using lines as style examples, whereas our TSB is applied at the word level. We used just a single word instance from the same author, as the style example. The target generation was set at word-level. We asked eight participants to rate generated images in two separate tasks. 

In the first task, we presented participants with randomly-ordered real and generated images, from our method and its baselines (SRNet~\cite{WuZLHLDB19} for scene text; Davis et al.~\cite{davis2020text} for handwriting). Users were asked to classify images as either {\em Original} (sampled from real datasets) or {\em Generated}. We present our findings in Table~\ref{tab:userStudy}. Numbers are the percent of times users considered an image to be real. Rows labeled as {\em Real} report the frequency of correctly identifying real photos. In both scene text and handwriting, users misidentified our results with genuine images {\em nearly twice as often} as the images generated by our baselines, testifying to the improved photo-realism of our results. 

The second task presented a source style image alongside photos with the same style but new content, generated by our method and SRNet~\cite{WuZLHLDB19} for scene text, and Davis et al.~\cite{davis2020text} for handwriting. Participants were asked to select the more realistic generated photo. The results, reported in the last row of Table~\ref{tab:userStudy}, show that the participants preferred our method over SRNet, 77.70\% of the times and over Davis et al.~\cite{davis2020text}, 77.65\% of the times. These results again testify to the heightened photo-realism of our method compared to existing work.



\section{Limitations of our approach}
\label{sec:failure}
\begin{figure}
    \centering
    \includegraphics[width=\linewidth]{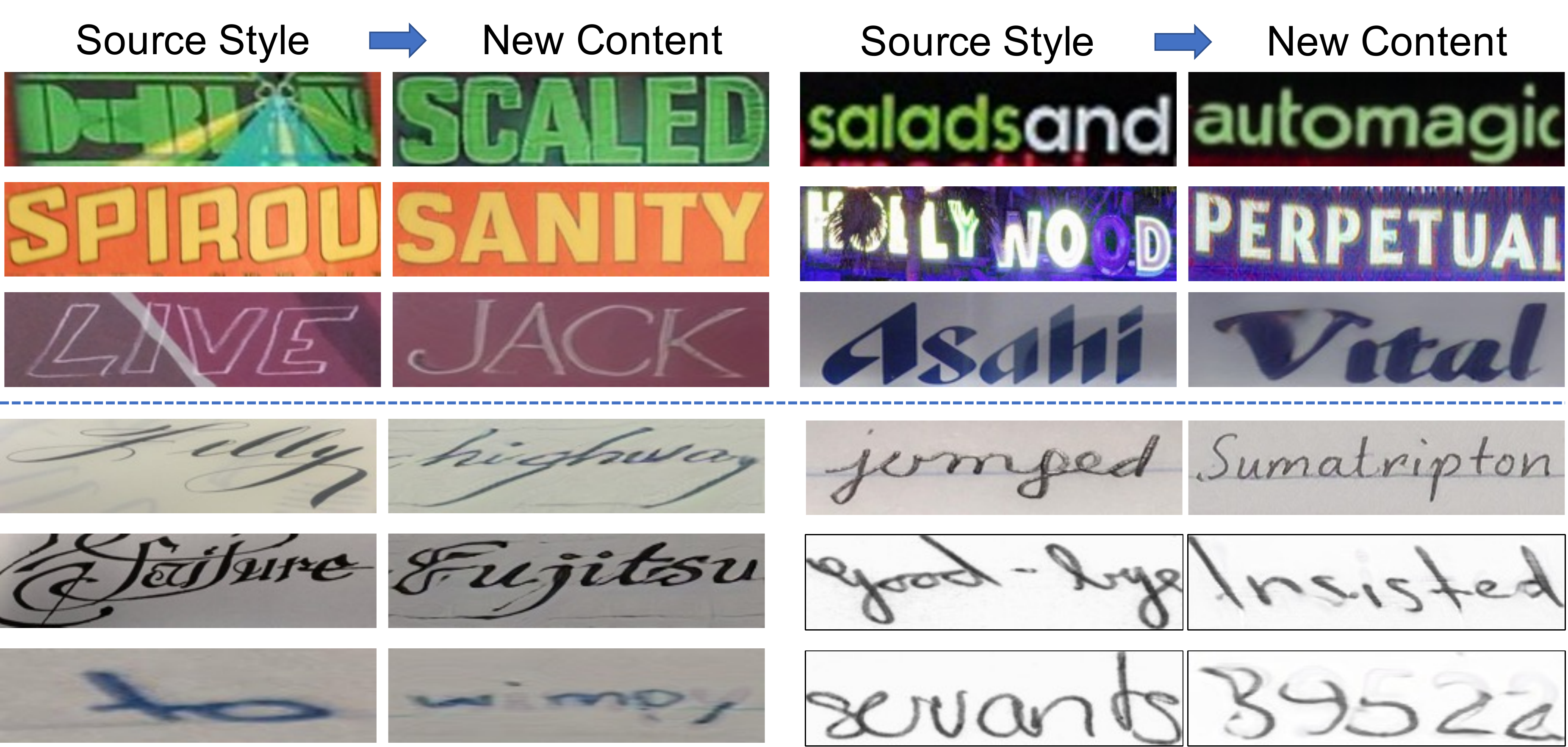}
    \caption{{\bf Limitations of our approach.} Qualitative example scene text (top) and handwritten (bottom) failures. These failures are due to local style variations (e.g., colors varying among characters), metallic colors which were not well represented in the training, and uniquely complex calligraphy for handwriting.}
    \label{fig:fail}
\end{figure}

Fig.~\ref{fig:fail} offers sample generation failures in scene text (top) and handwriting domains (bottom). Most of these examples share complex styles where: (a) The target foreground font color/style is inconsistent with the input style image, (b) Cases where the generation is not photo-realistic. Some of these complex scenario includes the text is written in metallic objects, different colors for different characters, etc. Regardless, in all these cases, our method managed to correctly generate the target content.

In case of handwriting, there are few instances where some of the characters in the target content are blurred or not generated in a realistic manner. This happens mostly in scenarios when the source style image is too short ($<3$ characters). Row 4 \& 5, Col. 1 presents failure cases where the input style is a very complex form of calligraphy. The other failure examples shown in the figure belong to the scenarios where the system didn't capture the cursive property of the input style or got the shear incorrectly. In our current set-up we used a pre-trained text recognizer which is trained only on scene text images. We believe that many such issues could be mitigated if we pre-trained model from handwritten domain itself. 

\section{Conclusion}
\label{sec:conc}
We present a method for disentangling real photos of text, extracting an opaque representation of all appearance aspects of the text and allowing transfer of this appearance to new content strings. To our knowledge, our method claims a number of novel capabilities. Most notably, our TSB requires only a single source style example: Style transfer is {\em one-shot}. Our disentanglement approach is further trained in a self-supervised manner, allowing the use of real photos for training, without style labels. Finally, unlike previous work we show synthetically-generated results on both scene text {\em and} handwritten text whereas existing methods were tailored to one domain or the other, but not both.  

Our method aims at use cases involving creative self expression and augmented reality (e.g., photo-realistic translation, leveraging multi-lingual OCR technologies~\cite{huang2021multiplexed}). Our method can be used for data generation and augmentation for training future OCR systems, as successfully done by others~\cite{GuptaVZ16,liao2020synthtext3d} and in other domains~\cite{masi2017rapid,masi2019face}. We are aware, however, that like other technologies, ours can be misused, possibly the same as deepfake faces can be used for misinformation. We see it as {\em imperative} that the abilities we describe are published, to facilitate research into detecting such misuse, e.g., by moving beyond fake faces to text, in benchmarks such as FaceForensics++~\cite{rossler2019ffpp} and the Deepfake Detection Challenge (DFDC)~\cite{dolhansky2019deepfake}. Our method can also be used to create training data for detecting fake text from images. Finally, we hope our work will encourage regulators and educators to address the inexorable rise of deepfake technologies.



\ifCLASSOPTIONcaptionsoff
  \newpage
\fi



%

\bibliographystyle{IEEEtran}
\bibliography{IEEEabrv,egbib}

\begin{thebibliography}{10}
\providecommand{\url}[1]{#1}
\csname url@samestyle\endcsname
\providecommand{\newblock}{\relax}
\providecommand{\bibinfo}[2]{#2}
\providecommand{\BIBentrySTDinterwordspacing}{\spaceskip=0pt\relax}
\providecommand{\BIBentryALTinterwordstretchfactor}{4}
\providecommand{\BIBentryALTinterwordspacing}{\spaceskip=\fontdimen2\font plus
\BIBentryALTinterwordstretchfactor\fontdimen3\font minus
  \fontdimen4\font\relax}
\providecommand{\BIBforeignlanguage}[2]{{%
\expandafter\ifx\csname l@#1\endcsname\relax
\typeout{** WARNING: IEEEtran.bst: No hyphenation pattern has been}%
\typeout{** loaded for the language `#1'. Using the pattern for}%
\typeout{** the default language instead.}%
\else
\language=\csname l@#1\endcsname
\fi
#2}}
\providecommand{\BIBdecl}{\relax}
\BIBdecl

\bibitem{bringhurst2004elements}
R.~Bringhurst, \emph{The elements of typographic style}.\hskip 1em plus 0.5em
  minus 0.4em\relax Hartley \& Marks Vancouver, 2004.

\bibitem{mediavilla1996calligraphy}
C.~Mediavilla, \emph{Calligraphy: from calligraphy to abstract painting}.\hskip
  1em plus 0.5em minus 0.4em\relax Scirpus Publ., 1996.

\bibitem{long2021scene}
S.~Long, X.~He, and C.~Yao, ``Scene text detection and recognition: The deep
  learning era,'' \emph{Int. J. Comput. Vis.}, vol. 129, no.~1, pp. 161--184,
  2021.

\bibitem{neumann2012real}
L.~Neumann and J.~Matas, ``Real-time scene text localization and recognition,''
  in \emph{IEEE Conf. Comput. Vis. Pattern Recog.}, 2012, pp. 3538--3545.

\bibitem{wang2011end}
K.~Wang, B.~Babenko, and S.~Belongie, ``End-to-end scene text recognition,'' in
  \emph{Int. Conf. Comput. Vis.}, 2011, pp. 1457--1464.

\bibitem{hassner2012computation}
T.~Hassner, M.~Rehbein, P.~A. Stokes, and L.~Wolf, ``Computation and
  palaeography: potentials and limits,'' \emph{Dagstuhl Reports}, vol.~2,
  no.~9, pp. 184--199, 2012.

\bibitem{hassner2014digital}
T.~Hassner, R.~Sablatnig, D.~Stutzmann, and S.~Tarte, ``Digital palaeography:
  New machines and old texts (dagstuhl seminar 14302),'' \emph{Dagstuhl
  Reports}, vol.~4, no.~7, 2014.

\bibitem{WuZLHLDB19}
L.~Wu, C.~Zhang, J.~Liu, J.~Han, J.~Liu, E.~Ding, and X.~Bai, ``Editing text in
  the wild,'' in \emph{Proc. Int. Conf. on Multimedia}, 2019, pp. 1500--1508.

\bibitem{YangHL20}
Q.~Yang, J.~Huang, and W.~Lin, ``{SwapText}: Image based texts transfer in
  scenes,'' in \emph{IEEE Conf. Comput. Vis. Pattern Recog.}, 2020, pp.
  14\,688--14\,697.

\bibitem{RoyBG020}
P.~Roy, S.~Bhattacharya, S.~Ghosh, and U.~Pal, ``{STEFANN:} scene text editor
  using font adaptive neural network,'' in \emph{IEEE Conf. Comput. Vis.
  Pattern Recog.}, 2020, pp. 13\,225--13\,234.

\bibitem{Yang0WG19}
S.~Yang, J.~Liu, W.~Wang, and Z.~Guo, ``{TET-GAN:} text effects transfer via
  stylization and destylization,'' in \emph{AAAI}, 2019, pp. 1238--1245.

\bibitem{AzadiFKWSD18}
S.~Azadi, M.~Fisher, V.~G. Kim, Z.~Wang, E.~Shechtman, and T.~Darrell,
  ``Multi-content {GAN} for few-shot font style transfer,'' in \emph{IEEE Conf.
  Comput. Vis. Pattern Recog.}, 2018, pp. 7564--7573.

\bibitem{FogelACML20}
S.~Fogel, H.~Averbuch{-}Elor, S.~Cohen, S.~Mazor, and R.~Litman, ``Scrabblegan:
  Semi-supervised varying length handwritten text generation,'' in \emph{IEEE
  Conf. Comput. Vis. Pattern Recog.}, 2020, pp. 4323--4332.

\bibitem{davis2020text}
B.~Davis, C.~Tensmeyer, B.~Price, C.~Wigington, B.~Morse, and R.~Jain, ``Text
  and style conditioned {GAN} for generation of offline handwriting lines,''
  \emph{arXiv preprint arXiv:2009.00678}, 2020.

\bibitem{Goodfellow14}
I.~J. Goodfellow, J.~Pouget-Abadie, M.~Mirza, B.~Xu, D.~Warde-Farley, S.~Ozair,
  A.~Courville, and Y.~Bengio, ``Generative adversarial nets,'' in \emph{Adv.
  Neural Inform. Process. Syst.}, 2014, pp. 2672–--2680.

\bibitem{Kingma14}
D.~P. Kingma and M.~Welling, ``Auto-encoding variational bayes,'' in \emph{Int.
  Conf. Learn. Represent.}, 2014.

\bibitem{Isola17}
P.~Isola, J.-Y. Zhu, T.~Zhou, and A.~A. Efros, ``Image-to-image translation
  with conditional adversarial networks,'' in \emph{IEEE Conf. Comput. Vis.
  Pattern Recog.}, 2017.

\bibitem{Zhu17}
J.-Y. Zhu, T.~Park, P.~Isola, and A.~A. Efros, ``Unpaired image-to-image
  translation using cycle-consistent adversarial networks,'' in \emph{Int.
  Conf. Comput. Vis.}, 2017.

\bibitem{GatysEB16}
L.~A. Gatys, A.~S. Ecker, and M.~Bethge, ``Image style transfer using
  convolutional neural networks,'' in \emph{IEEE Conf. Comput. Vis. Pattern
  Recog.}, 2016, pp. 2414--2423.

\bibitem{Johnson16}
J.~Johnson, A.~Alahi, and L.~Fei-Fei, ``Perceptual losses for real-time style
  transfer and super-resolution,'' in \emph{Eur. Conf. Comput. Vis.}, 2016.

\bibitem{Huang17}
X.~Huang and S.~Belongie, ``Arbitrary style transfer in real-time with adaptive
  instance normalization,'' in \emph{Int. Conf. Comput. Vis.}, 2017.

\bibitem{HuangLBK18}
X.~Huang, M.~Liu, S.~J. Belongie, and J.~Kautz, ``Multimodal unsupervised
  image-to-image translation,'' in \emph{Eur. Conf. Comput. Vis.}, ser. Lecture
  Notes in Computer Science, vol. 11207, 2018, pp. 179--196.

\bibitem{liu2019few}
M.-Y. Liu, X.~Huang, A.~Mallya, T.~Karras, T.~Aila, J.~Lehtinen, and J.~Kautz,
  ``Few-shot unsupervised image-to-image translation,'' in \emph{Int. Conf.
  Comput. Vis.}, 2019, pp. 10\,551--10\,560.

\bibitem{choi2020stargan}
Y.~Choi, Y.~Uh, J.~Yoo, and J.-W. Ha, ``{StarGAN} v2: Diverse image synthesis
  for multiple domains,'' in \emph{IEEE Conf. Comput. Vis. Pattern Recog.},
  2020, pp. 8188--8197.

\bibitem{kang2020ganwriting}
L.~Kang, P.~Riba, Y.~Wang, M.~Rusi{\~n}ol, A.~Forn{\'e}s, and M.~Villegas,
  ``{GANwriting}: Content-conditioned generation of styled handwritten word
  images,'' in \emph{Eur. Conf. Comput. Vis.}, 2020, pp. 273--289.

\bibitem{Karras18}
T.~Karras, T.~Aila, S.~Laine, and J.~Lehtinen, ``Progressive growing of {GANs}
  for improved quality, stability, and variation,'' in \emph{Int. Conf. Learn.
  Represent.}, 2018.

\bibitem{KarrasLA19}
T.~Karras, S.~Laine, and T.~Aila, ``A style-based generator architecture for
  generative adversarial networks,'' in \emph{IEEE Conf. Comput. Vis. Pattern
  Recog.}, 2019, pp. 4401--4410.

\bibitem{KarrasLAHLA20}
T.~Karras, S.~Laine, M.~Aittala, J.~Hellsten, J.~Lehtinen, and T.~Aila,
  ``Analyzing and improving the image quality of {StyleGAN},'' in \emph{IEEE
  Conf. Comput. Vis. Pattern Recog.}, 2020, pp. 8107--8116.

\bibitem{YangLLG17}
S.~Yang, J.~Liu, Z.~Lian, and Z.~Guo, ``Awesome typography: Statistics-based
  text effects transfer,'' in \emph{IEEE Conf. Comput. Vis. Pattern Recog.},
  2017, pp. 7464--7473.

\bibitem{li2020fet}
W.~Li, Y.~He, Y.~Qi, Z.~Li, and Y.~Tang, ``{FET-GAN}: Font and effect transfer
  via k-shot adaptive instance normalization,'' in \emph{AAAI}, vol.~34,
  no.~02, 2020, pp. 1717--1724.

\bibitem{ZhanZL19}
F.~Zhan, H.~Zhu, and S.~Lu, ``Spatial fusion {GAN} for image synthesis,'' in
  \emph{IEEE Conf. Comput. Vis. Pattern Recog.}, 2019, pp. 3653--3662.

\bibitem{GomezBGGKR19}
R.~Gomez, A.~F. Biten, L.~G{\'{o}}mez, J.~Gibert, D.~Karatzas, and
  M.~Rusi{\~{n}}ol, ``Selective style transfer for text,'' in \emph{Int. Conf.
  Document Analysis and Recog.}, 2019, pp. 805--812.

\bibitem{ChangZPM18}
B.~Chang, Q.~Zhang, S.~Pan, and L.~Meng, ``Generating handwritten chinese
  characters using {CycleGAN},'' in \emph{Winter Conf. on App. of Comput.
  Vision}, 2018, pp. 199--207.

\bibitem{AlonsoMM19}
E.~Alonso, B.~Moysset, and R.~O. Messina, ``Adversarial generation of
  handwritten text images conditioned on sequences,'' in \emph{Int. Conf.
  Document Analysis and Recog.}, 2019, pp. 481--486.

\bibitem{HeZRS16}
K.~He, X.~Zhang, S.~Ren, and J.~Sun, ``Deep residual learning for image
  recognition,'' in \emph{IEEE Conf. Comput. Vis. Pattern Recog.}, 2016, pp.
  770--778.

\bibitem{he2017mask}
K.~He, G.~Gkioxari, P.~Doll{\'a}r, and R.~Girshick, ``Mask {R-CNN},'' in
  \emph{Int. Conf. Comput. Vis.}, 2017, pp. 2961--2969.

\bibitem{samara2004typography}
T.~Samara, \emph{Typography workbook: A real-world guide to using type in
  graphic design}.\hskip 1em plus 0.5em minus 0.4em\relax Rockport Publishers,
  2004.

\bibitem{hood1992pen}
M.~Hood, ``Pen, ink, \& evidence: A study of writing and writing materials for
  the penman, collector, and document detective,'' 1992.

\bibitem{SimonyanZ14a}
K.~Simonyan and A.~Zisserman, ``Very deep convolutional networks for
  large-scale image recognition,'' in \emph{Int. Conf. Learn. Represent.},
  Y.~Bengio and Y.~LeCun, Eds., 2015.

\bibitem{BaekKLPHYOL19}
J.~Baek, G.~Kim, J.~Lee, S.~Park, D.~Han, S.~Yun, S.~J. Oh, and H.~Lee, ``What
  is wrong with scene text recognition model comparisons? dataset and model
  analysis,'' in \emph{Int. Conf. Comput. Vis.}, 2019, pp. 4714--4722.

\bibitem{liao2020mask}
M.~Liao, G.~Pang, J.~Huang, T.~Hassner, and X.~Bai, ``Mask {TextSpotter} v3:
  Segmentation proposal network for robust scene text spotting,'' in \emph{Eur.
  Conf. Comput. Vis.}, 2020.

\bibitem{textocr}
A.~Singh, G.~Pang, M.~Toh, J.~Huang, T.~Hassner, and W.~Galuba, ``{TextOCR}:
  Towards large-scale end-to-end reasoning for arbitrary-shaped scene text,''
  in \emph{IEEE Conf. Comput. Vis. Pattern Recog.}, 2021.

\bibitem{rath2007word}
T.~M. Rath and R.~Manmatha, ``Word spotting for historical documents,''
  \emph{Int. J. Doc. Anal. Recog.}, vol.~9, no. 2-4, pp. 139--152, 2007.

\bibitem{kleber2013cvl}
F.~Kleber, S.~Fiel, M.~Diem, and R.~Sablatnig, ``{CVL}-database: An off-line
  database for writer retrieval, writer identification and word spotting,'' in
  \emph{Int. Conf. Document Analysis and Recog.}, 2013, pp. 560--564.

\bibitem{sanchez2014icfhr2014}
J.~A. S{\'a}nchez, V.~Romero, A.~H. Toselli, and E.~Vidal, ``{ICFHR2014}
  competition on handwritten text recognition on transcriptorium datasets
  ({HTRtS}),'' in \emph{Int. Conf. Front. Hand. Recog.}, 2014, pp. 785--790.

\bibitem{MartiB02}
U.-V. Marti and H.~Bunke, ``The {IAM}-database: an english sentence database
  for offline handwriting recognition,'' \emph{Int. J. Doc. Anal. Recog.},
  vol.~5, no.~1, pp. 39--46, 2002.

\bibitem{ChengBXZPZ17}
Z.~Cheng, F.~Bai, Y.~Xu, G.~Zheng, S.~Pu, and S.~Zhou, ``Focusing attention:
  Towards accurate text recognition in natural images,'' in \emph{Int. Conf.
  Comput. Vis.}, 2017, pp. 5086--5094.

\bibitem{pytorch}
``Pytorch distributed framework,'' available online: \url{https://pytorch.org}.

\bibitem{GuptaVZ16}
A.~Gupta, A.~Vedaldi, and A.~Zisserman, ``Synthetic data for text localisation
  in natural images,'' in \emph{IEEE Conf. Comput. Vis. Pattern Recog.}, 2016,
  pp. 2315--2324.

\bibitem{KaratzasSUIBMMMAH13}
D.~Karatzas, F.~Shafait, S.~Uchida, M.~Iwamura, L.~G. i~Bigorda, S.~R. Mestre,
  J.~Mas, D.~F. Mota, J.~Almaz{\'{a}}n, and L.~de~las Heras, ``{ICDAR} 2013
  robust reading competition,'' in \emph{Int. Conf. Document Analysis and
  Recog.}, 2013, pp. 1484--1493.

\bibitem{KaratzasGNGBIMN15}
D.~Karatzas, L.~Gomez{-}Bigorda, A.~Nicolaou, S.~K. Ghosh, A.~D. Bagdanov,
  M.~Iwamura, J.~Matas, L.~Neumann, V.~R. Chandrasekhar, S.~Lu, F.~Shafait,
  S.~Uchida, and E.~Valveny, ``{ICDAR} 2015 competition on robust reading,'' in
  \emph{Int. Conf. Document Analysis and Recog.}, 2015, pp. 1156--1160.

\bibitem{singh2019towards}
A.~Singh, V.~Natarjan, M.~Shah, Y.~Jiang, X.~Chen, D.~Parikh, and M.~Rohrbach,
  ``Towards {VQA} models that can read,'' in \emph{IEEE Conf. Comput. Vis.
  Pattern Recog.}, 2019, pp. 8317--8326.

\bibitem{OpenImages2}
I.~Krasin, T.~Duerig, N.~Alldrin, V.~Ferrari, S.~Abu-El-Haija, A.~Kuznetsova,
  H.~Rom, J.~Uijlings, S.~Popov, S.~Kamali, M.~Malloci, J.~Pont-Tuset, A.~Veit,
  S.~Belongie, V.~Gomes, A.~Gupta, C.~Sun, G.~Chechik, D.~Cai, Z.~Feng,
  D.~Narayanan, and K.~Murphy, ``{OpenImages}: A public dataset for large-scale
  multi-label and multi-class image classification.'' \emph{Dataset available
  from https://storage.googleapis.com/openimages/web/index.html}, 2017.

\bibitem{WangBSS04}
Z.~Wang, A.~C. Bovik, H.~R. Sheikh, and E.~P. Simoncelli, ``Image quality
  assessment: from error visibility to structural similarity,'' \emph{IEEE
  Trans. Image Process.}, vol.~13, no.~4, pp. 600--612, 2004.

\bibitem{HeuselRUNH17}
M.~Heusel, H.~Ramsauer, T.~Unterthiner, B.~Nessler, and S.~Hochreiter, ``{GANs}
  trained by a two time-scale update rule converge to a local nash
  equilibrium,'' in \emph{Adv. Neural Inform. Process. Syst.}, 2017, pp.
  6626--6637.

\bibitem{KhrulkovO18}
V.~Khrulkov and I.~V. Oseledets, ``Geometry score: {A} method for comparing
  generative adversarial networks,'' in \emph{Int. Conf. Mach. Learning}, J.~G.
  Dy and A.~Krause, Eds., vol.~80, 2018, pp. 2626--2634.

\bibitem{perez2003poisson}
P.~P{\'e}rez, M.~Gangnet, and A.~Blake, ``Poisson image editing,'' in \emph{ACM
  Trans. Graph.}, 2003, pp. 313--318.

\bibitem{srnetCode}
``Tensorflow implementation for editing text in the wild,'' available online:
  \url{https://github.com/youdao-ai/SRNet}.

\bibitem{huang2021multiplexed}
J.~Huang, G.~Pang, R.~Kovvuri, M.~Toh, K.~J. Liang, P.~Krishnan, X.~Yin, and
  T.~Hassner, ``A multiplexed network for end-to-end, multilingual {OCR},'' in
  \emph{IEEE Conf. Comput. Vis. Pattern Recog.}, 2021.

\bibitem{liao2020synthtext3d}
M.~Liao, B.~Song, S.~Long, M.~He, C.~Yao, and X.~Bai, ``{SynthText3D}:
  synthesizing scene text images from {3D} virtual worlds,'' \emph{Science
  China Information Sciences}, vol.~63, no.~2, pp. 1--14, 2020.

\bibitem{masi2017rapid}
I.~Masi, T.~Hassner, A.~T. Tran, and G.~Medioni, ``Rapid synthesis of massive
  face sets for improved face recognition,'' in \emph{Int. Conf. on Automatic
  Face and Gesture Recognition}, 2017, pp. 604--611.

\bibitem{masi2019face}
I.~Masi, A.~T. Tran, T.~Hassner, G.~Sahin, and G.~Medioni, ``Face-specific data
  augmentation for unconstrained face recognition,'' \emph{Int. J. Comput.
  Vis.}, vol. 127, no. 6-7, pp. 642--667, 2019.

\bibitem{rossler2019ffpp}
A.~Rossler, D.~Cozzolino, L.~Verdoliva, C.~Riess, J.~Thies, and M.~Nie{\ss}ner,
  ``Faceforensics++: Learning to detect manipulated facial images,'' in
  \emph{Int. Conf. Comput. Vis.}, 2019, pp. 1--11.

\bibitem{dolhansky2019deepfake}
B.~Dolhansky, R.~Howes, B.~Pflaum, N.~Baram, and C.~C. Ferrer, ``The deepfake
  detection challenge ({DFDC}) preview dataset,'' \emph{arXiv preprint
  arXiv:1910.08854}, 2019.

\end{thebibliography}


\begin{thebibliography}{10}
\providecommand{\url}[1]{#1}
\csname url@samestyle\endcsname
\providecommand{\newblock}{\relax}
\providecommand{\bibinfo}[2]{#2}
\providecommand{\BIBentrySTDinterwordspacing}{\spaceskip=0pt\relax}
\providecommand{\BIBentryALTinterwordstretchfactor}{4}
\providecommand{\BIBentryALTinterwordspacing}{\spaceskip=\fontdimen2\font plus
\BIBentryALTinterwordstretchfactor\fontdimen3\font minus
  \fontdimen4\font\relax}
\providecommand{\BIBforeignlanguage}[2]{{%
\expandafter\ifx\csname l@#1\endcsname\relax
\typeout{** WARNING: IEEEtran.bst: No hyphenation pattern has been}%
\typeout{** loaded for the language `#1'. Using the pattern for}%
\typeout{** the default language instead.}%
\else
\language=\csname l@#1\endcsname
\fi
#2}}
\providecommand{\BIBdecl}{\relax}
\BIBdecl

\bibitem{he2017mask}
K.~He, G.~Gkioxari, P.~Doll{\'a}r, and R.~Girshick, ``Mask {R-CNN},'' in
  \emph{Int. Conf. Comput. Vis.}, 2017, pp. 2961--2969.

\bibitem{KarrasLAHLA20}
T.~Karras, S.~Laine, M.~Aittala, J.~Hellsten, J.~Lehtinen, and T.~Aila,
  ``Analyzing and improving the image quality of {StyleGAN},'' in \emph{IEEE
  Conf. Comput. Vis. Pattern Recog.}, 2020, pp. 8107--8116.

\bibitem{BaekKLPHYOL19}
J.~Baek, G.~Kim, J.~Lee, S.~Park, D.~Han, S.~Yun, S.~J. Oh, and H.~Lee, ``What
  is wrong with scene text recognition model comparisons? dataset and model
  analysis,'' in \emph{Int. Conf. Comput. Vis.}, 2019, pp. 4714--4722.

\bibitem{SimonyanZ14a}
K.~Simonyan and A.~Zisserman, ``Very deep convolutional networks for
  large-scale image recognition,'' in \emph{Int. Conf. Learn. Represent.},
  Y.~Bengio and Y.~LeCun, Eds., 2015.

\bibitem{WuZLHLDB19}
L.~Wu, C.~Zhang, J.~Liu, J.~Han, J.~Liu, E.~Ding, and X.~Bai, ``Editing text in
  the wild,'' in \emph{Proc. Int. Conf. on Multimedia}, 2019, pp. 1500--1508.

\bibitem{srnetDatagen}
``Srnet-datagen,'' available online:
  \url{https://github.com/youdao-ai/SRNet-Datagen}.

\bibitem{synthtext}
``Synthtext,'' available online: \url{https://github.com/ankush-me/SynthText}.

\bibitem{unrealtext}
``Unrealtext,'' available online: \url{https://github.com/Jyouhou/UnrealText}.

\bibitem{perez2003poisson}
P.~P{\'e}rez, M.~Gangnet, and A.~Blake, ``Poisson image editing,'' in \emph{ACM
  Trans. Graph.}, 2003, pp. 313--318.

\bibitem{KaratzasSUIBMMMAH13}
D.~Karatzas, F.~Shafait, S.~Uchida, M.~Iwamura, L.~G. i~Bigorda, S.~R. Mestre,
  J.~Mas, D.~F. Mota, J.~Almaz{\'{a}}n, and L.~de~las Heras, ``{ICDAR} 2013
  robust reading competition,'' in \emph{Int. Conf. Document Analysis and
  Recog.}, 2013, pp. 1484--1493.

\bibitem{singh2019towards}
A.~Singh, V.~Natarjan, M.~Shah, Y.~Jiang, X.~Chen, D.~Parikh, and M.~Rohrbach,
  ``Towards vqa models that can read,'' in \emph{IEEE Conf. Comput. Vis.
  Pattern Recog.}, 2019, pp. 8317--8326.

\bibitem{MartiB02}
U.-V. Marti and H.~Bunke, ``The iam-database: an english sentence database for
  offline handwriting recognition,'' \emph{Int. J. Doc. Anal. Recog.}, vol.~5,
  no.~1, pp. 39--46, 2002.

\end{thebibliography}

%








\end{document}


\title{TextStyleBrush: Transfer of Text Aesthetics from a Single Example\\--Supplemental Material--}

\author{Praveen Krishnan,
        Rama Kovvuri,
        Guan Pang,
        Boris Vassilev,
        Tal Hassner\\
        Facebook AI
        
{\small 
\{pkrishnan,ramakovvuri,gpang,borisva,thassner\}@fb.com}
}

\markboth{Journal of \LaTeX\ Class Files,~Vol.~14, No.~8, August~2015}%
{Shell \MakeLowercase{\textit{et al.}}: Bare Demo of IEEEtran.cls for Computer Society Journals}
\maketitle

\section{Introduction}
\label{sec:intro}
This supplemental provides details of the underlying networks used as part of our proposed Text Style Brush (TSB) architecture (Sec.~\ref{sec:netArch}) and details of the synthetic dataset (Sec.~\ref{sec:dataset}). Finally we offer multitudes of qualitative results from different datasets and domains mentioned in the paper, with examples of results stitched back into the original photos and word level results, all in Sec.~\ref{sec:qual}.

\section{Network details}
\label{sec:netArch}
Our proposed TSB architecture consists of seven networks: \emph{style encoder}, $(F_s)$, \emph{content encoder}, $(F_c)$, style mapping network, $M$, \emph{stylized text generator}, $(G)$, and networks $C,R$, and $D$ used to compute our various loss functions, representing \emph{typeface classifier, recognizer}, and \emph{discriminator}, respectively. The following notations are used to describe each component of the network: convolutional layer (Conv), pooling layer (Pool), residual convolutional block (ResBlock), average pooling layer (AvgPool), fully connected layer (FC), stride ($s$), upsampling factor ($up$), kernel size ($k$) and channels ($c$).

Table~\ref{tab:encStyleArch} and~\ref{tab:encContArch} presents the encoder architecture for style and content respectively. Note that we use similar architecture for both encoders except the first and last layer. The style encoder takes input a localized scene RGB image along  with  word  bounding  box (Word BB). The last layer uses  a  region  of  interest (RoI) align  operator  as  described  in  Mask  R-CNN~\cite{he2017mask} which  pools features coming from the desired word region. The style representation is produced by $e_s=F_s(\mathbb{I}_{s,c})$, a 512D representation.

The content encoder, shown in Table~\ref{tab:encContArch} takes in a gray scale image. While training the input image is set to fixed dimensions $64\times256$ and the output feature matrix is of dimension $512 \times 4\times 16$. Here, $512$ denotes the number of channels. While testing the input is set to variable width $64\times W$, which allows target strings of different lengths to be generated in original aspect ratio. The content representation is given by $e_c=F_c(\mathcal{I}_{\hat{s},c})$. 

Table~\ref{tab:styleArch} presents the style mapping network architecture. It takes the style encoded vector, $e_s$, as input and outputs layer specific style components denoted as $w_{s,i}, \forall i \in [1,15]$. The Norm refers to the Normalization layer which is given as:
\begin{equation}
    \frac{e_s}{\sqrt{\frac{1}{512}\sum_{i=1}^{512}e_{s,i}^2}+\epsilon}.
\end{equation}
Here, $\epsilon=1\mathrm{e}{-8}$ is used for numerical stability.

Table~\ref{tab:stgArch}, presents the stylized text generator architecture. Our generator is a modified version of the StyleGAN2 with skip connections and without progressive growing. The input to the generator is the learned content feature matrix, $e_c$, of dimension $512 \times 4 \times 16$. Each block shown in the table contains modulated convolutional~\cite{KarrasLAHLA20} style layers (StyleConv), RGB layers (RGBConv), and soft mask layers (MaskConv). These layers (Style, RGB, Mask) uses the corresponding style vectors denoted as $w_{s,i}$ for modulating the weights. The output from our generator is taken from the last RGB layer which produces an image of dimension $64 \times 256$ along with a soft mask image of same dimensions. We base our discriminator architecture on StyleGAN2~\cite{KarrasLAHLA20} where the input is of dimension $3\times 64\times 256$ and the output is a scalar value denoting the score for real/fake.

We used the standard off-the-shelf architectures for our text recognizer, $R$, and typeface classifier, $C$. The recognizer uses the existing pre-trained word recognition model of Baek et al.~\cite{BaekKLPHYOL19}. Of the models described in that paper, we chose the one with the following configuration, though we did not optimize for this choice: (1) spatial transformation network (STN) using thin-plate spline (TPS) transformation, (2) feature extraction using ResNet network, (3) sequence modelling using BiLSTM, and (4) an attention-based sequence prediction. Similarly, we built our typeface classifier using the VGG19 network~\cite{SimonyanZ14a}.

\begin{table}[!h]
    \centering
    \begin{tabular}{|c|ccc|c|}\hline
        \textbf{Layers} & \multicolumn{3}{c|}{\textbf{Configurations}} & \textbf{Output} \\\hline
        Input & \multicolumn{3}{c|}{RGB Image ($\mathbb{I}_{s,c}$), Word BB} & $256 \times256$\\\hline
        Conv0-1 & $c:32$ && $k:3\times 3$ & $256 \times256$\\\hline
        Conv0-2 & $c:64$ && $k:3\times 3$ & $256 \times256$\\\hline
        Pool1 && $s:2$ & $k:2\times 2$ & $128 \times128$\\\hline
        ResBlock1 
        & \multicolumn{3}{c|}{$
        \begin{bmatrix}
            c :128, k :3 \times 3\\ 
            c :128, k :3 \times 3
        \end{bmatrix} \times 1$}
    & $128 \times128$\\\hline
    Conv1 & $c:128$ && $k:3\times 3$ & $128 \times128$\\\hline
    Pool2 && $s:2$ & $k:2\times 2$ & $64 \times64$\\\hline
    ResBlock2 
        & \multicolumn{3}{c|}{$
        \begin{bmatrix}
            c :256, k :3 \times 3\\ 
            c :256, k :3 \times 3
        \end{bmatrix} \times 2$}
    & $64 \times64$\\\hline
    Conv2 & $c:256$ && $k:3\times 3$ & $64 \times64$\\\hline
    Pool3 && $s:2$ & $k:2\times 2$ & $32 \times32$\\\hline
    ResBlock3 
        & \multicolumn{3}{c|}{$
        \begin{bmatrix}
            c :512, k :3 \times 3\\ 
            c :512, k :3 \times 3
        \end{bmatrix} \times 5$}
    & $32 \times32$\\\hline
    Conv3 & $c:512$ && $k:3\times 3$ & $32 \times32$\\\hline
    Pool4 && $s:2$ & $k:2\times 2$ & $16 \times16$\\\hline
    ResBlock4 
        & \multicolumn{3}{c|}{$
        \begin{bmatrix}
            c :512, k :3 \times 3\\ 
            c :512, k :3 \times 3
        \end{bmatrix} \times 3$}
    & $16 \times16$\\\hline
    Conv4-1 & $c:512$ &s:1& $k:3\times 3$ & $16 \times16$\\ \hline
    
    ROI Align & $c:512$ &&& $1 \times1$\\ \hline
    \end{tabular}
    \caption{Style encoder architecture.}
    \label{tab:encStyleArch}
\end{table}

\begin{table}[!h]
    \centering
    \begin{tabular}{|c|ccc|c|}\hline
        \textbf{Layers} & \multicolumn{3}{c|}{\textbf{Configurations}} & \textbf{Output} \\\hline
        Input & \multicolumn{3}{c|}{Gray Scale Image ($\mathcal{I}_{\hat{s},c}$)} & $64 \times256$\\\hline
        Conv0-1 & $c:32$ && $k:3\times 3$ & $64 \times256$\\\hline
        Conv0-2 & $c:64$ && $k:3\times 3$ & $64 \times256$\\\hline
        Pool1 && $s:2$ & $k:2\times 2$ & $32 \times128$\\\hline
        ResBlock1 
        & \multicolumn{3}{c|}{$
        \begin{bmatrix}
            c :128, k :3 \times 3\\ 
            c :128, k :3 \times 3
        \end{bmatrix} \times 1$}
    & $32 \times128$\\\hline
    Conv1 & $c:128$ && $k:3\times 3$ & $32 \times128$\\\hline
    Pool2 && $s:2$ & $k:2\times 2$ & $16 \times64$\\\hline
    ResBlock2 
        & \multicolumn{3}{c|}{$
        \begin{bmatrix}
            c :256, k :3 \times 3\\ 
            c :256, k :3 \times 3
        \end{bmatrix} \times 2$}
    & $16 \times64$\\\hline
    Conv2 & $c:256$ && $k:3\times 3$ & $16 \times64$\\\hline
    Pool3 && $s:2$ & $k:2\times 2$ & $8 \times32$\\\hline
    ResBlock3 
        & \multicolumn{3}{c|}{$
        \begin{bmatrix}
            c :512, k :3 \times 3\\ 
            c :512, k :3 \times 3
        \end{bmatrix} \times 5$}
    & $8 \times32$\\\hline
    Conv3 & $c:512$ && $k:3\times 3$ & $8 \times32$\\\hline
    Pool4 && $s:2$ & $k:2\times 2$ & $4 \times16$\\\hline
    ResBlock4 
        & \multicolumn{3}{c|}{$
        \begin{bmatrix}
            c :512, k :3 \times 3\\ 
            c :512, k :3 \times 3
        \end{bmatrix} \times 3$}
    & $4 \times16$\\\hline
    Conv4-1 & $c:512$ &s:1& $k:3\times 3$ & $4 \times16$\\\hline
    \end{tabular}
    \caption{Content encoder architecture.}
    \label{tab:encContArch}
\end{table}

\begin{table}[!h]
    \centering
    \begin{tabular}{|c|ccc|c|}\hline
        \textbf{Layers} & \multicolumn{3}{c|}{\textbf{Configurations}} & \textbf{Output} \\\hline
        Input & \multicolumn{3}{c|}{Style Vector ($e_s$)} & $1 \times512$\\\hline
        Norm. &&&& $1 \times512$\\\hline
        FC1 &$c:512$&&$k:1\times1$& $1 \times512$\\\hline
        FC2 &$c:7680$&&$k:1\times1$& $15 \times512$\\\hline
        \end{tabular}
    \caption{Style mapping network architecture.}
    \label{tab:styleArch}
\end{table}

\begin{table}[!h]
    \centering
    \resizebox{\columnwidth}{!}{%
    \begin{tabular}{|c|ccc|c|}\hline
        \textbf{Layers} & \multicolumn{3}{c|}{\textbf{Configurations}} & \textbf{Output} \\\hline
        Input & \multicolumn{3}{c|}{Content Feature Matrix ($e_c$)} & $4 \times16$\\\hline
        Block1 
        & \multicolumn{3}{c|}{$
        \begin{bmatrix}
            \text{StyleConv} & w_{s,1} & up=1\\
            \text{RGBConv} & w_{s,2} & up=1\\ 
            \text{MaskConv} & w_{s,3} & up=1\\ 
        \end{bmatrix}$}
    & $4\times16$\\\hline
        Block2 
        & \multicolumn{3}{c|}{$
        \begin{bmatrix}
            \text{StyleConv} & w_{s,3} & up=2\\ 
            \text{StyleConv} & w_{s,4} & up=1\\ 
            \text{RGBConv} & w_{s,5} & up=1\\ 
            \text{MaskConv} & w_{s,6} & up=1\\ 
        \end{bmatrix}$}
    & $8\times32$\\\hline
    Block3 
        & \multicolumn{3}{c|}{$
        \begin{bmatrix}
            \text{StyleConv} & w_{s,6} & up=2\\ 
            \text{StyleConv} & w_{s,7} & up=1\\ 
            \text{RGBConv} & w_{s,8} & up=1\\ 
            \text{MaskConv} & w_{s,9} & up=1\\ 
        \end{bmatrix}$}
    & $16\times64$\\\hline
    Block4 
        & \multicolumn{3}{c|}{$
        \begin{bmatrix}
            \text{StyleConv} & w_{s,9} & up=2\\ 
            \text{StyleConv} & w_{s,10} & up=1\\ 
            \text{RGBConv} & w_{s,11} & up=1\\ 
            \text{MaskConv} & w_{s,12} & up=1\\ 
        \end{bmatrix}$}
    & $32\times128$\\\hline
    Block5 
        & \multicolumn{3}{c|}{$
        \begin{bmatrix}
            \text{StyleConv} & w_{s,12} & up=2\\ 
            \text{StyleConv} & w_{s,13} & up=1\\ 
            \text{RGBConv} & w_{s,14} & up=1\\
            \text{MaskConv} & w_{s,15} & up=1\\ 
        \end{bmatrix}$}
    & $64\times256$\\\hline
    \end{tabular}}
    \caption{Stylized text generator architecture. Here each $\text{StyleConv}$ block uses 512 output channels and kernel size, $k=3\times3$. While the $\text{RGBConv},\text{MaskConv}$ uses three and one output channels respectively with a kernel size, $k=1\times1$.}
    \label{tab:stgArch}
\end{table}


\section{Dataset Details}
\label{sec:dataset}

\begin{figure}[t]
    \centering
    \includegraphics[width=\linewidth]{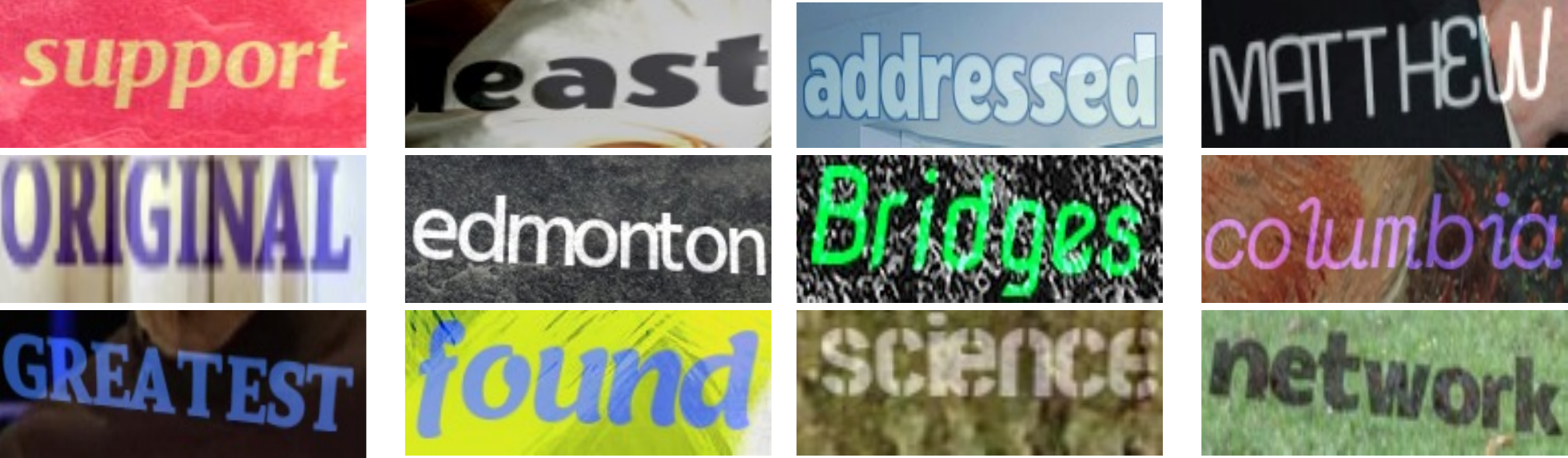}
    \caption{Sample word images from our synthetic datasets.}
    \label{fig:synthImages}
\end{figure}

\subsection{Synthetic dataset}
\label{subsec:synthDataset}
The Synth-Paired and the Synth-Font datasets are created following the pipeline presented by SRNet~\cite{WuZLHLDB19}. We used a third party implementation of the rendering pipeline~\cite{srnetDatagen}. The background images were downloaded from the original code repository of SynthText~\cite{synthtext}. We used $\sim$2K different font styles which were downloaded from the UnrealText repository available at~\cite{unrealtext}. Fig.~\ref{fig:synthImages} offers some sample synthetic images from these datasets.

\section{Qualitative Results}
\label{sec:qual}
\begin{figure*}
    \centering
    \includegraphics[width=\linewidth]{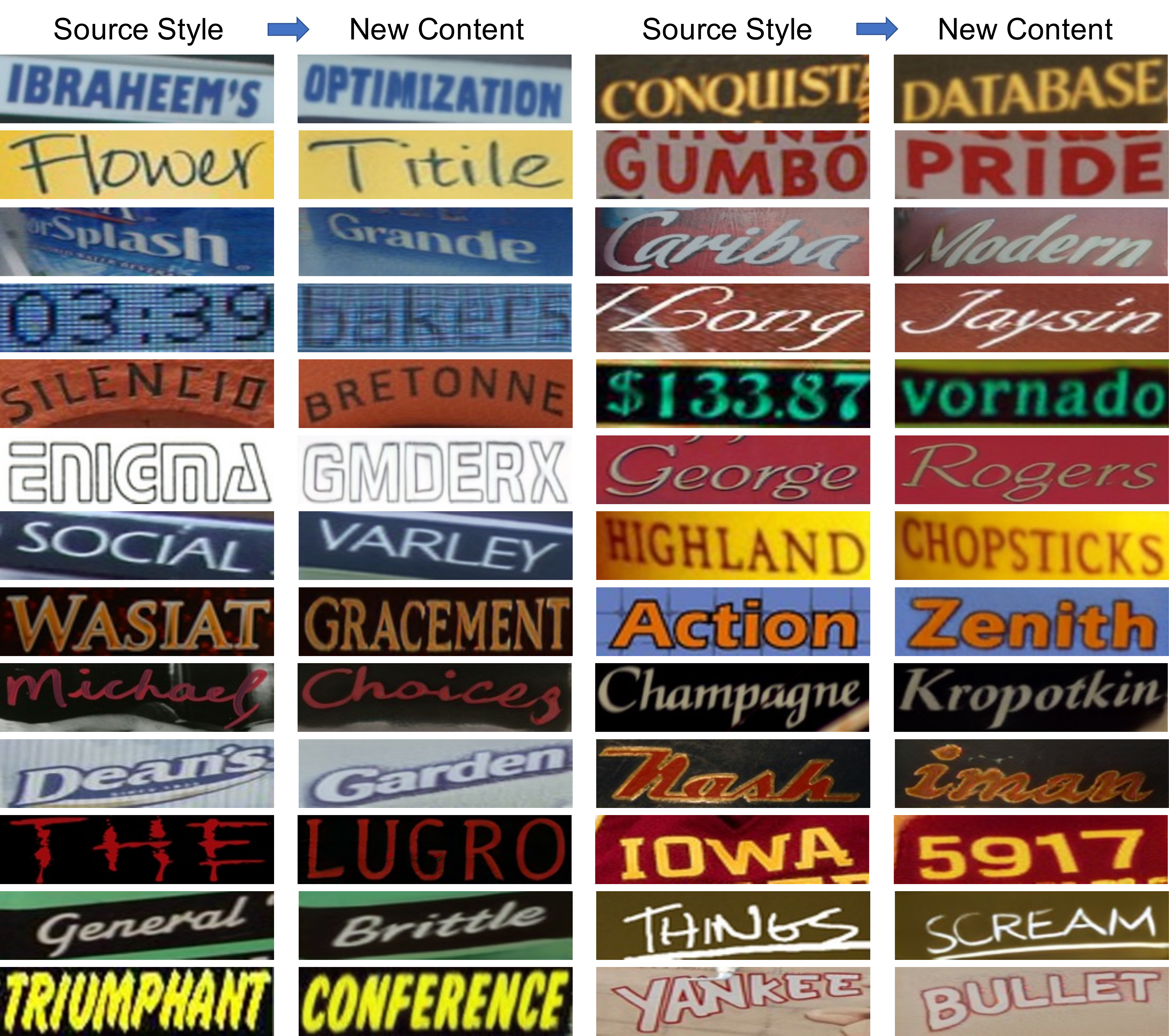}
    \caption{Word level style transfer results: TextVQA~\cite{singh2019towards}}
    \label{fig:textvqa_words}
\end{figure*}

\begin{figure*}
    \centering
    \includegraphics[width=0.8\linewidth]{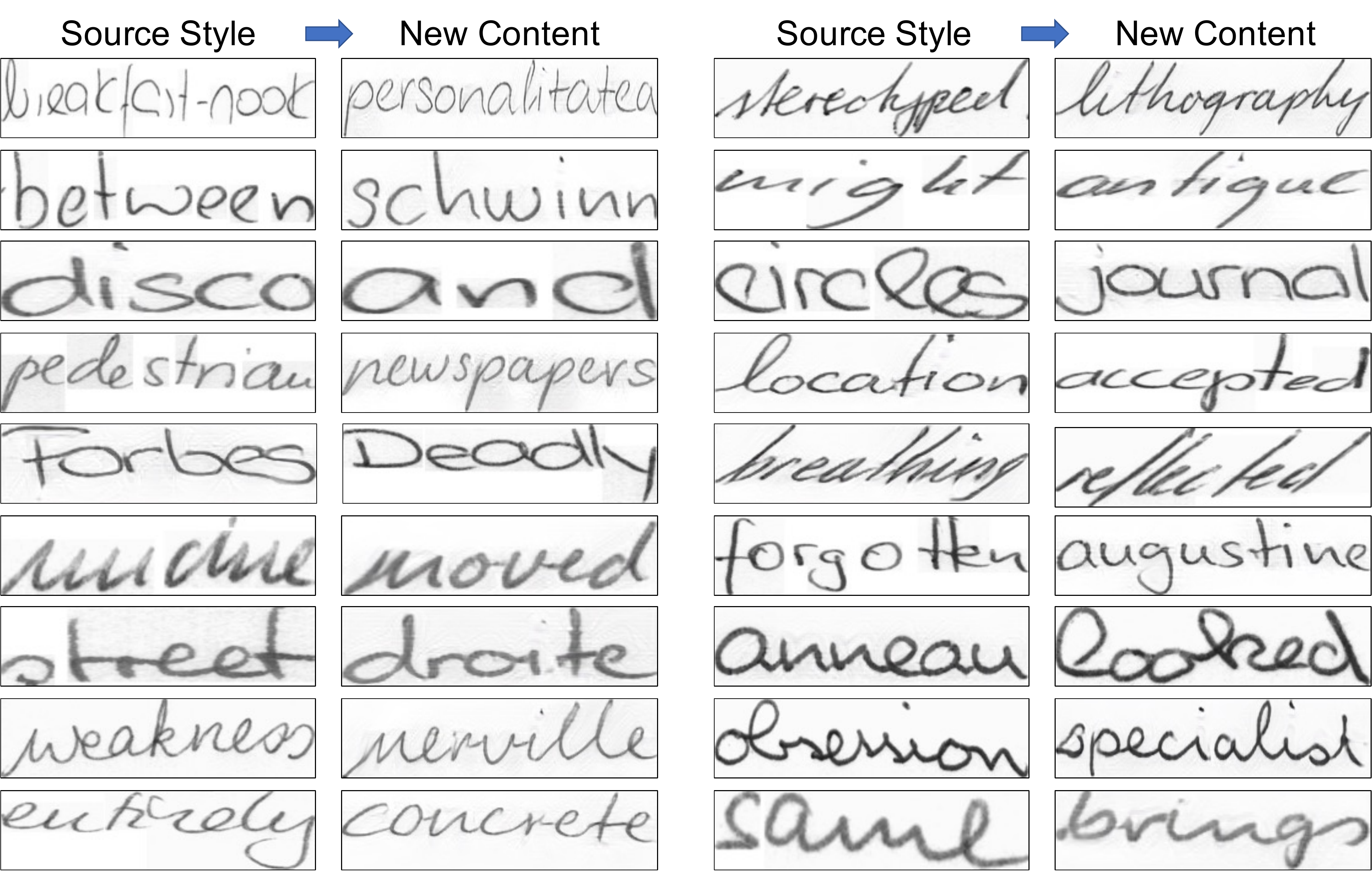}
    \caption{Word level style transfer results: IAM~\cite{MartiB02}}
    \label{fig:iam_words}
\end{figure*}

\begin{figure*}
    \centering
    \includegraphics[width=0.8\linewidth]{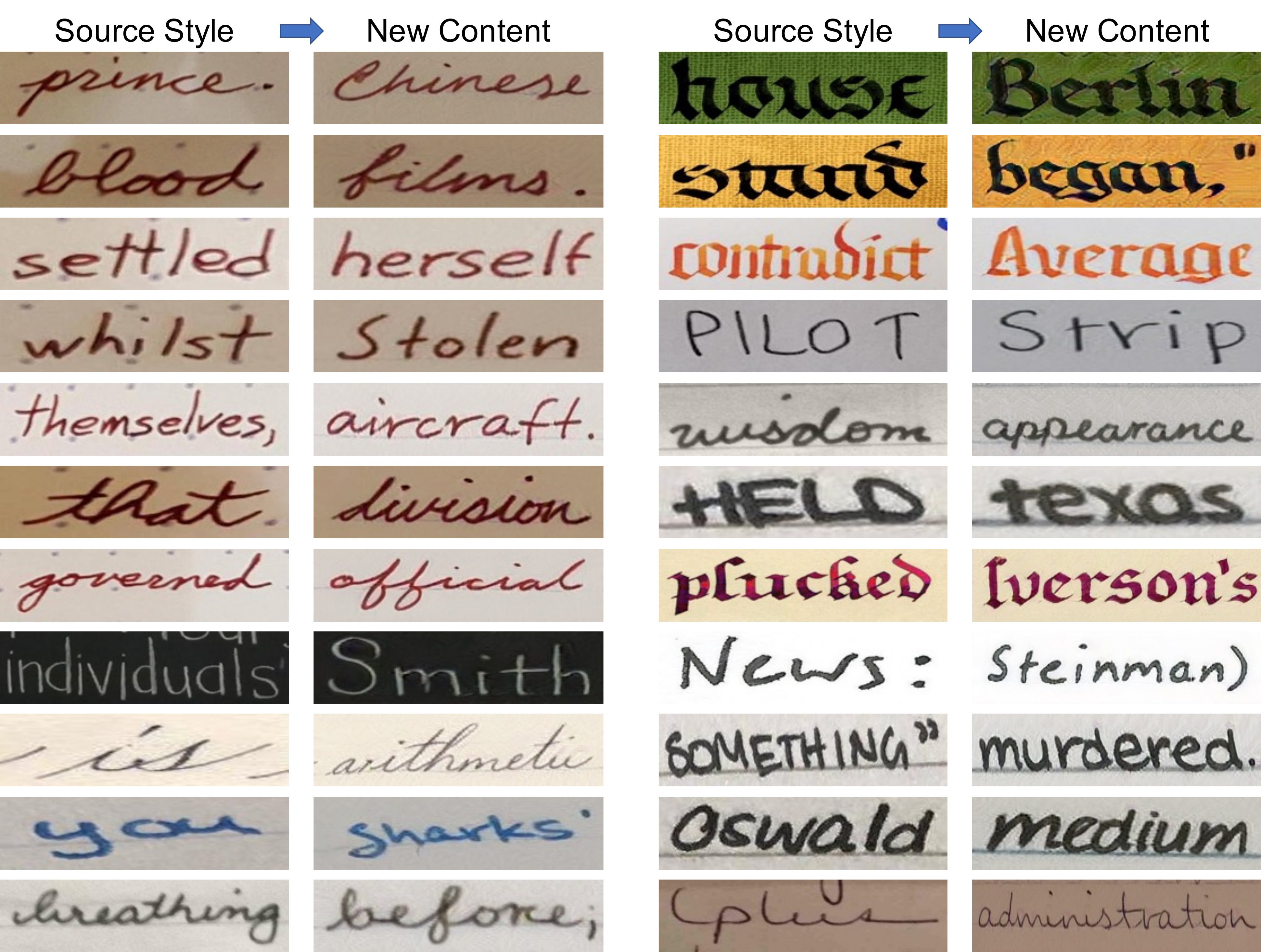}
    \caption{Word level style transfer results: Imgur5K}
    \label{fig:imgur_words}
\end{figure*}

We offer many qualitative style transfer results from all the datasets used in this work. Word (box) level style transfer results are shown in the following figures: ICDAR13 (Fig~\ref{fig:icdar13_words}), TextVQA (Fig~\ref{fig:textvqa_words}), IAM Handwriting database (Fig~\ref{fig:iam_words}), Imgur5K (Fig~\ref{fig:imgur_words}). 

\begin{figure*}
    \centering
    \includegraphics[width=0.9\linewidth]{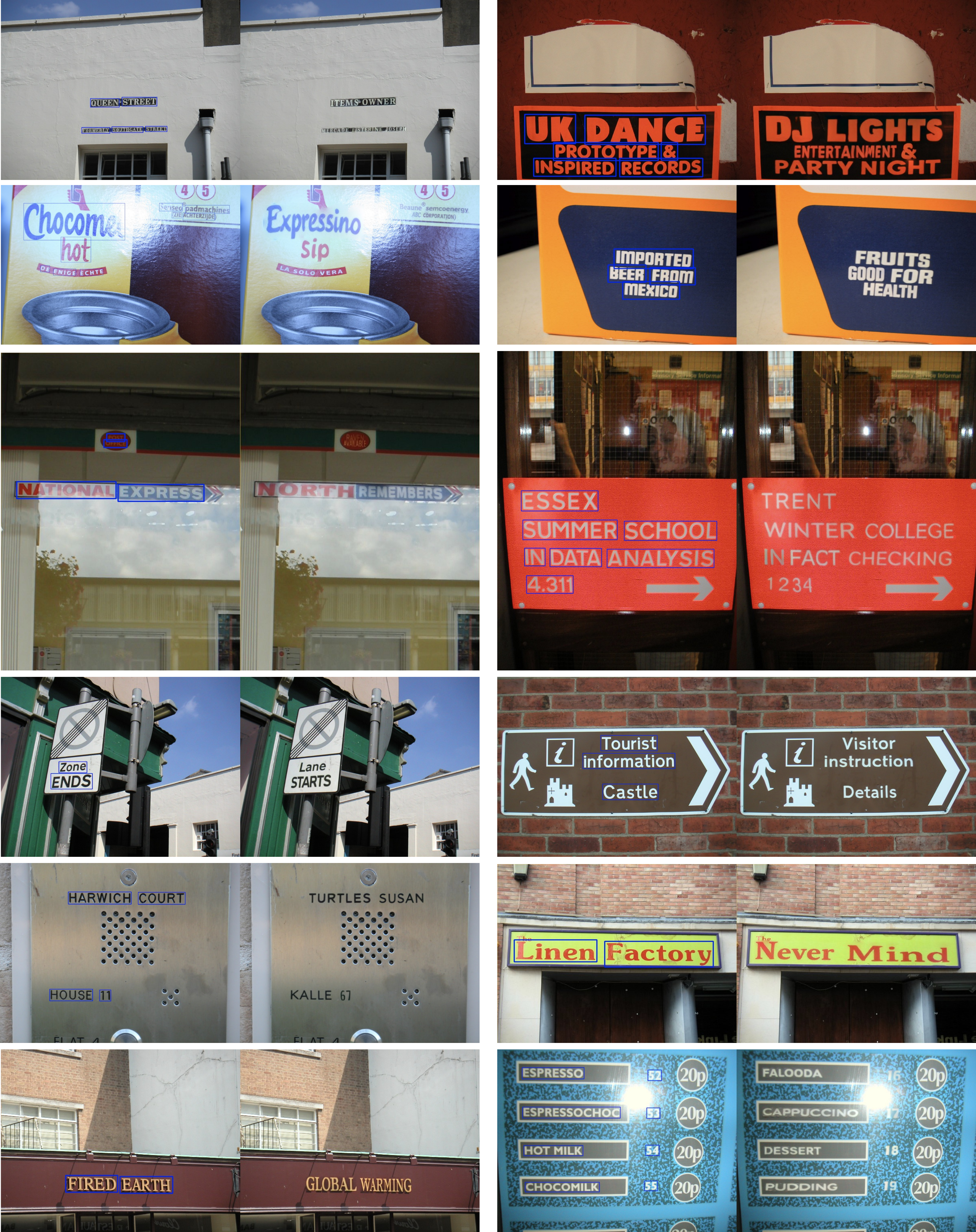}
    \caption{Scene text editing results on ICDAR13~\cite{KaratzasSUIBMMMAH13} dataset. On left, we show the original image, and on right we present the edited image with selected words (shown in blue boxes) replaced with a new content following the same style of original content.}
    \label{fig:stitch_real_icdar13}
\end{figure*}

\begin{figure*}
    \centering
    \includegraphics[width=0.9\linewidth]{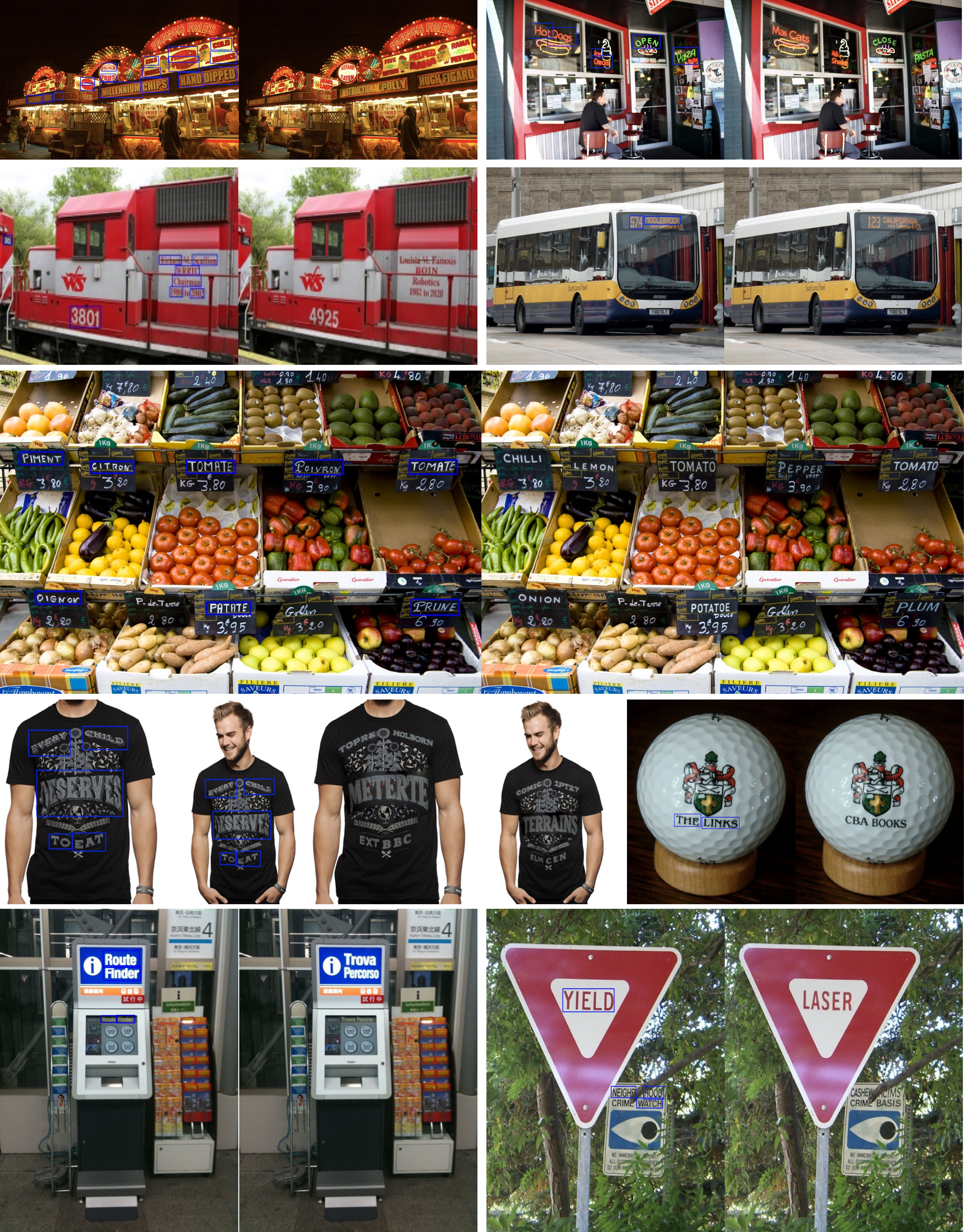}
    \caption{Scene text editing results on TextVQA~\cite{singh2019towards} dataset. On left, we show the original image, and on right we present the edited image with  selected words (shown in blue boxes) replaced with a new content following the same style of original content.}
    \label{fig:stitch_real_textvqa}
\end{figure*}

In addition, we provide sample text editing results where, given an input photo with an automatically detected text box, our method replaces the contents of that text box with a new string, stylized to appear similar to the original. The new content is then stitched back into the photo, entirely replacing the contents of the detected box. The new content is stitched into the photo using simple Poisson blending~\cite{perez2003poisson}. Fig.~\ref{fig:stitch_real_icdar13},~\ref{fig:stitch_real_textvqa} presents few samples of input photos along with their edited versions, from images taken from the ICDAR13~\cite{KaratzasSUIBMMMAH13} and TextVQA set~\cite{singh2019towards} respectively. 



\begin{figure*}
    \centering
    \includegraphics[width=\linewidth]{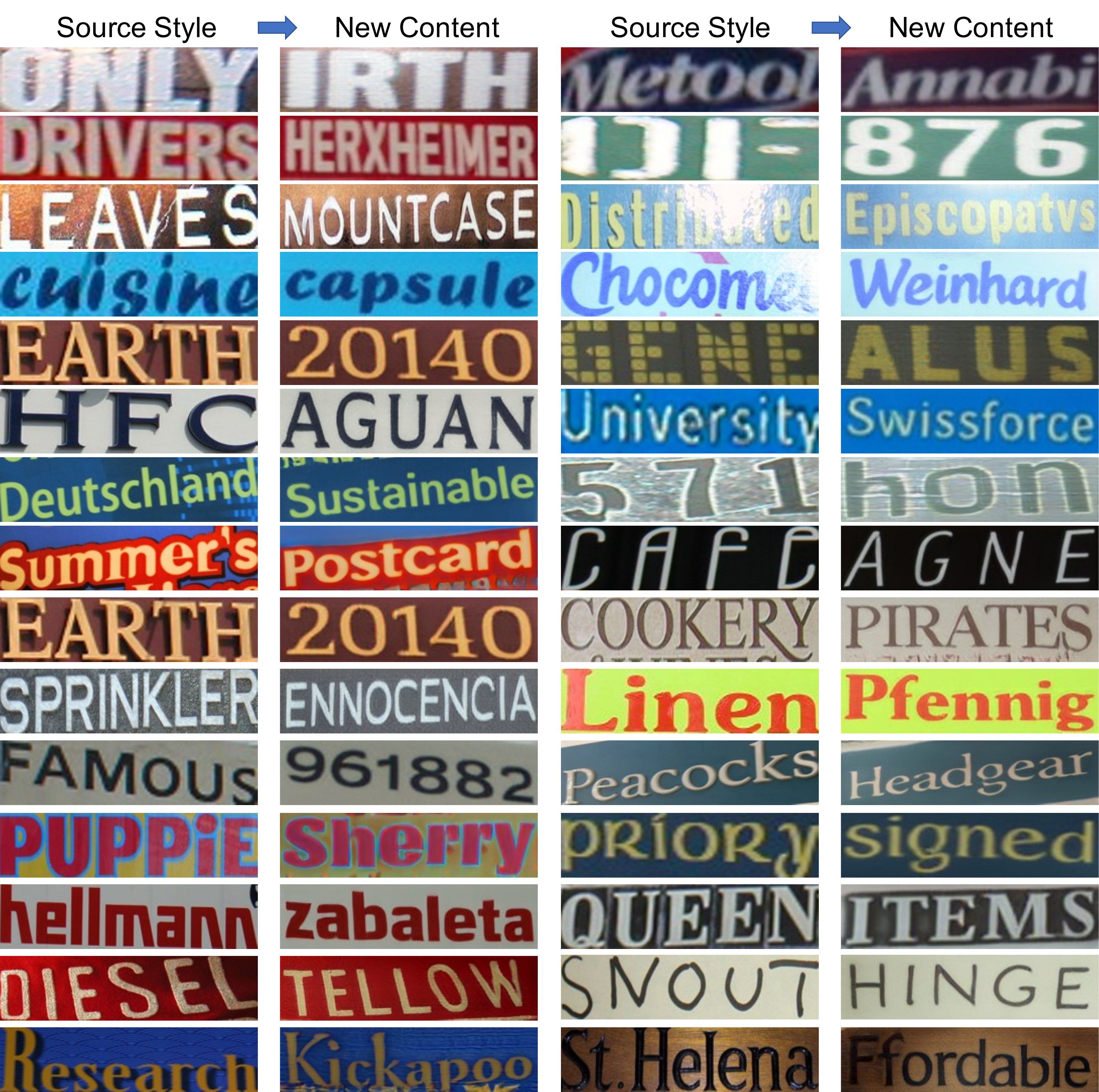}
    \caption{Word level style transfer results: ICDAR13~\cite{KaratzasSUIBMMMAH13}}
    \label{fig:icdar13_words}
\end{figure*}

\bibliographystyle{IEEEtran}
\bibliography{IEEEabrv,egbib_supp}